\documentclass[10pt,twocolumn,letterpaper]{article}

\usepackage{cvpr}              
\definecolor{cvprblue}{rgb}{0.21,0.49,0.74}
\usepackage[pagebackref,breaklinks,colorlinks,allcolors=cvprblue]{hyperref}
\usepackage{booktabs}
\usepackage{tabularx}
\usepackage{listings} 
\usepackage{xcolor}

\usepackage[accsupp]{axessibility}

\definecolor{codebg}{RGB}{248,248,248}
\definecolor{codeframe}{RGB}{220,220,220}
\definecolor{codecomment}{RGB}{90,90,90}

\lstdefinestyle{custom_style}{
  backgroundcolor=\color{codebg},
  frame=single,
  rulecolor=\color{codeframe},
  basicstyle=\ttfamily\footnotesize,
  breaklines=true,
  breakatwhitespace=true,
  columns=fullflexible,
  keepspaces=true,
  showstringspaces=false,
  upquote=true,
  tabsize=2,
  xleftmargin=4pt,
  xrightmargin=4pt
}

\usepackage[T1]{fontenc}
\usepackage[utf8]{inputenc} 
\usepackage{textcomp}
\def\paperID{EgoVis-9} 
\def\confName{CVPR}
\def\confYear{2026}

\title{How to Correctly Make Mistakes: A Framework for Constructing and Benchmarking Mistake Aware Egocentric Procedural Videos}

\author{Olga Loginova\\
University of Trento\\
Italy\\
{\tt\small olga.loginova@unitn.it}
\and
Frank Keller\\
University of Edinburgh\\
United Kingdom\\
{\tt\small keller@inf.ed.ac.uk}
}

\begin{document}
\maketitle
\begin{abstract}
Reliable procedural monitoring in video requires exposure to naturally occurring human errors and the recoveries that follow. In egocentric recordings, mistakes are often partially occluded by hands and revealed through subtle object state changes, while existing procedural datasets provide limited and inconsistent mistake and correction traces. We present PIE-V (Psychologically Inspired Error injection for Videos), a framework for constructing and benchmarking mistake-aware egocentric procedural videos by augmenting clean keystep procedures with controlled, human-plausible deviations. PIE-V combines a psychology-informed error planner conditioned on procedure phase and semantic step load, a correction planner that models recovery behavior, an LLM writer that performs cascade-consistent rewrites, and an LLM judge that validates procedural coherence and repairs failures. For video segment edits, PIE-V synthesizes replacement clips with text-guided video generation and stitches them into the episode to preserve visual plausibility. Applied to 17 tasks and 50 Ego-Exo4D scenarios, PIE-V injects 102 mistakes and generates 27 recovery corrections. For benchmarking, we introduce a unified taxonomy and a human rubric with nine metrics that cover step-level and procedure-level quality, including plausibility, procedure logic with annotator confidence, state change coherence, and grounding between text and video. Using this protocol, we audit several existing resources and compare PIE-V against a freeform LLM generation baseline under the same criteria.
Together, the framework and rubric support post-completion verification for egocentric procedural mistake detection and correction.
\end{abstract}    
\section{Introduction}
\label{sec:intro}

To err is human; to err \textbf{humanly plausible} is hard.
Procedural assistants that watch, guide, or evaluate stepwise activities can fail if trained only on ideal executions.
In real kitchens, workshops, and labs, people omit steps, swap substeps, use the wrong tool, or execute a step slightly off.
Learning and evaluating such behavior therefore requires datasets with realistic deviations, yet these are hard to collect and standardize at scale, leaving current resources sparse and heterogeneous~\cite{bacharidis2025visionbasedmistakeanalysisprocedural}.
Such datasets are needed to train mistake detectors and to evaluate state tracking, consequential error localization, and recovery after an error, not only segment-level anomaly flags~\cite{Flaborea2024PREGOOMA,guo2025proceduralmistakedetectionaction,storks2025transparentcoherentproceduralmistake}.
They also support benchmarking of multimodal procedural assistants for egocentric guidance, post hoc procedure verification, and error-aware tutoring, where subtle mistakes and corrections matter beyond the final action label~\cite{haneji2025egooopsdatasetmistakeaction,peddi2024captaincook4ddatasetunderstandingerrors}.

\begin{figure*}[t]
    \centering
    \includegraphics[width=0.7\textwidth]{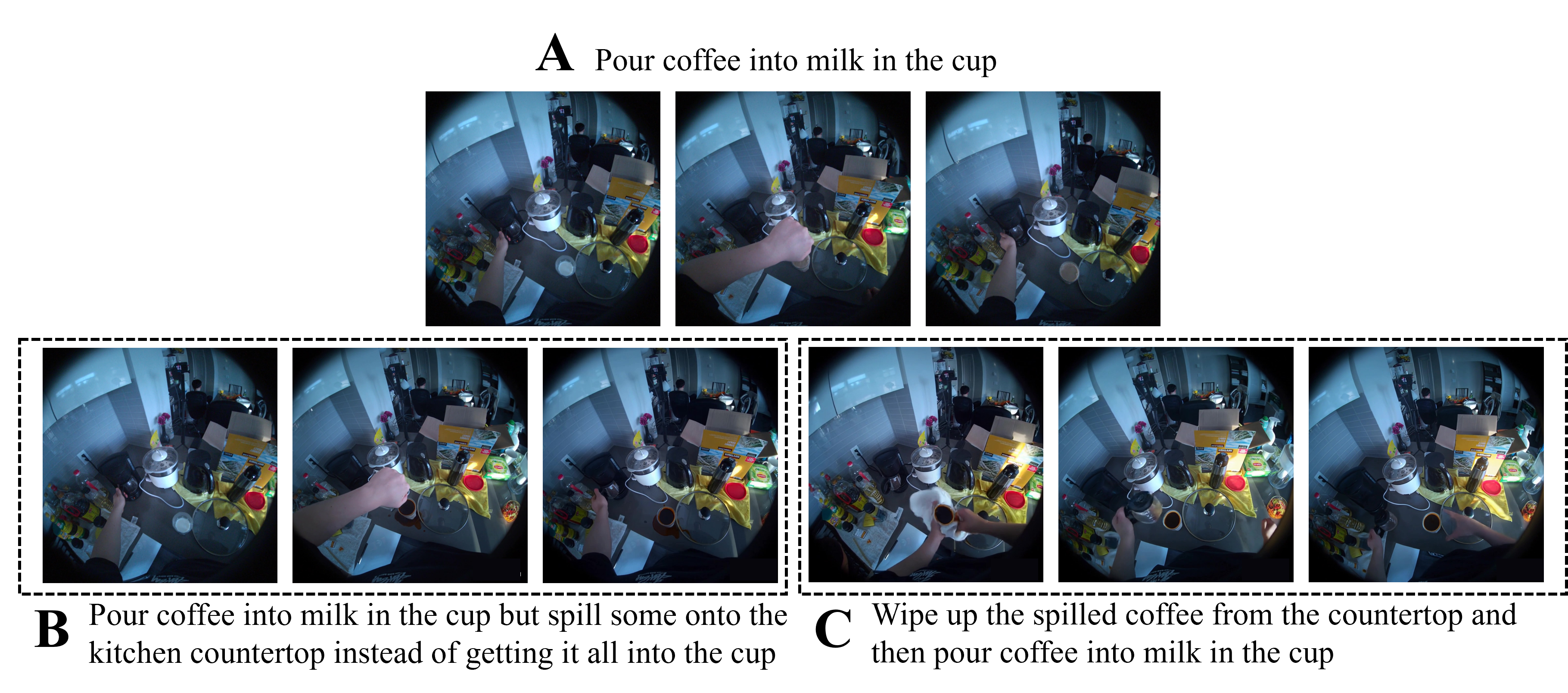}
    \caption{PIE-V example on an Ego-Exo4D step from ``Making Coffee Latte'': (A) reference step; (B) wrong execution with an observable spill; (C) correction that restores procedural consistency by cleaning and redoing the pour.}
    \label{fig:piev_coffee_example}
\end{figure*}

The community is moving toward more structured mistake reasoning.
Current approaches model errors through action effects and state changes~\cite{guo2025proceduralmistakedetectionaction, Flaborea2024PREGOOMA} and emphasize coherence~\cite{storks2025transparentcoherentproceduralmistake}.
Staged mistakes scale better~\cite{zhong2023learningprocedureawarevideorepresentation,peddi2024captaincook4ddatasetunderstandingerrors}, but naive perturbations often violate preconditions, create impossible object states, or break the task's causal structure~\cite{li2025mistakeattributionfinegrainedmistake, narasimhan2023learningverificationtaskstructure}.
This makes mistake resources difficult to compare across domains and often too underspecified to support robust recovery modeling.

How can we correctly make mistakes in procedural videos to provide mistake detection models with quality data?
A useful mistake-enriched dataset should satisfy two requirements: (1) the procedure remains executable and logically coherent as a sequence; (2) injected mistakes resemble human error patterns rather than arbitrary corruption.
We operationalize this with \textbf{PIE-V}~\footnote{Code and demos: \href{https://github.com/ologin/PIE-V/releases/tag/cvprw26}{https://github.com/ologin/PIE-V}.} (\textbf{P}sychologically \textbf{I}nspired \textbf{E}rror injection for \textbf{V}ideos), a scalable pipeline for augmenting procedural video datasets with mistake-aware variants.
PIE-V instantiates five universal error types (Deletion, Insertion, Transposition, Substitution, Wrong Execution) and controls when errors occur by procedure phase.
For Substitution and Wrong Execution, we use semantic roles within each step to localize the error and control its severity through role importance and step load. 
PIE-V also generates realistic Corrections, since natural errors often trigger recovery such as redoing a step, inserting a missing action, or undoing an incorrect state~\cite{Tamborello2013ALM}.

\Cref{fig:piev_coffee_example} illustrates a typical PIE-V error and correction trace on a real procedural step. PIE-V modifies both the instruction sequence and the corresponding video segment to maintain episode-level coherence.

A second challenge is evaluation.
Existing procedural mistake datasets are valuable resources, but they were typically designed for objectives such as detecting visual deviations from a reference step or procedure~\cite{Ding2023EveryMCA,haneji2025egooopsdatasetmistakeaction,Ghoddoosian2023WeaklySupervisedAS}.
For post-completion mistake detection and correction, where the goal is to validate the full procedure trace after a task appears complete, this yields only partial coverage. Annotations may capture visually salient deviations that are informative for their original tasks, but they do not necessarily correspond to consequential execution errors at the level of the full procedure. PIE-V avoids such visual ``artifacts'' by using video generation to create more subtle, behaviorally credible deviations.
We propose a rubric-style evaluation for mistake-aware procedural video datasets and use it to audit existing resources and compare generation strategies.
The rubric scores whether an event is a consequential execution error, whether its effects remain procedurally consistent, and whether the sample provides enough structure to study recovery.

Our contributions are four-fold: (1) a unified taxonomy of procedural errors and semantic roles; (2) a multi-criteria human rubric for mistake assessment; (3) an extensive audit comparing PIE-V against SoTA LLM baselines and existing datasets; and (4) PIE-V, a psychology-informed pipeline for semantics-aware mistake injection.

\begin{table*}[t]
  \caption{Qualitative cognitive motivations behind each error type in our taxonomy and their phase tendencies.}
  \label{tab:taxonomy}
  \centering
  \small
  \setlength{\tabcolsep}{6pt}
  \renewcommand{\arraystretch}{1.08}
  \begin{tabularx}{\textwidth}{@{}p{0.18\textwidth}X@{}}
    \toprule
    \textbf{Error type} & \textbf{Key psychological interpretation and phase tendency} \\
    \midrule
    Substitution &
    Confusion between similar steps, associative interference, or schema competition; tends to be rarer in later phases once the execution pattern stabilizes. \\
    Wrong execution &
    Execution slips and capture errors (local parameter mistakes), often due to unfamiliarity and motor learning early on; can occur throughout and may reappear toward the end under fatigue. \\
    Deletion &
    Lapses due to memory overload and post-completion vulnerability; tends to increase over the procedure, often peaking toward the end as attention drops. \\
    Insertion &
    Overgeneralization or associative activation (including unintended repetition); generally rarer, but can increase in the middle when multiple routines overlap and the performer is in the flow. \\
    Transposition &
    Sequencing and planning slips under high cognitive load; can occur under time pressure or fatigue, often when step ordering constraints are weak or when attention is divided. \\
    \bottomrule
  \end{tabularx}
\end{table*}
\section{Universal taxonomy of mistakes}
\label{sec:tax}

Our taxonomy draws on three sources: (i) existing mistake datasets and their annotation practices, (ii) sequence edit operations as a transferable structural view of procedural deviations, and (iii) cognitive error theory~\cite{Reason_1990}, which distinguishes plan-level changes from execution-level deviations and links plausibility to where an error occurs and how it disrupts task state.

We define five mistake types and treat \textsc{Correction} (\textsc{C}) as a separate component. Correction is reactive: it is triggered by the error to return the executor to the reference trace. 

The five mistake types are:
\textsc{Deletion} (\textsc{D}), a required step is missing;
\textsc{Insertion} (\textsc{I}), an extra step is added;
\textsc{Substitution} (\textsc{S}), an intended step is replaced by a different step;
\textsc{Transposition} (\textsc{T}), steps are executed in the wrong order;
and \textsc{Wrong Execution} (\textsc{WE}), the intended step occurs but with incorrect local parameters.
The first four types are structural edits on the step sequence and align with classical edit operations. Deletion, Insertion, and Substitution correspond to Levenshtein style edits~\cite{Levenshtein1965BinaryCC}. Transposition covers sequencing errors, including adjacent swaps characteristic of Damerau style edits~\cite{Damerau1964ATF}. Wrong Execution is related to Substitution but captures execution level deviations that preserve the step identity while changing how it is carried out. This type is closer to slips and lapses in cognitive accounts of human error.

Wrong Execution and many instances of Substitution require localization within a step. We therefore represent steps through \textit{semantic roles}~\cite{gildea-jurafsky-2002-automatic,palmer-etal-2005-proposition} and apply edits to targeted role arguments such as Object, Coobject, Instrument, Location, Destination, Origin, Purpose, and Manner. Role importance provides an explicit notion of severity by distinguishing high impact arguments (Object, Coobject\footnote{Coobject labels a secondary argument beyond the primary Object, often the target/recipient (e.g., in APPLY/FIT) or the second item being combined (e.g., in MIX/ADD).}
) from medium impact roles (Location, Destination, Origin, Instrument) and low impact modifiers (Manner, Temporal, Degree, Quantity). This role-based view is consistent with work that localizes mistakes inside steps and analyzes their attribution, and with evidence that procedural understanding depends on recovering and using step arguments, including implicit ones~\cite{li2025mistakeattributionfinegrainedmistake,batra-etal-2025-predicting}.

\Cref{tab:taxonomy} summarizes the taxonomy and qualitative cognitive motivations. In later sections we use this unified language to map heterogeneous datasets onto comparable categories. When a dataset uses task-specific labels that do not align cleanly with these types, we treat the mapping as approximate.

\section{Rubric for mistake-aware dataset assessment}
\label{sec:rubric}

\begin{table*}[t]
  \caption{Overview of the PIE-V dataset assessment rubric.}
  \label{tab:rubric_overview}
  \centering
  \small
  \setlength{\tabcolsep}{6pt}
  \renewcommand{\arraystretch}{1.08}
  \begin{tabularx}{\textwidth}{@{}p{0.28\textwidth}p{0.18\textwidth}X@{}}
    \toprule
    \textbf{Metric} & \textbf{Scale} & \textbf{What it measures} \\
    \midrule
    Error Validity &
    Binary &
    Whether the deviation should be treated as a mistake under the procedure-level benchmark, rather than a benign variation. \\
    Human Plausibility &
    Likert (1--5) &
    How natural the mistake appears in context, avoiding both overly perfect staging and implausible corruption. \\
    Confusability &
    Likert (1--5) &
    How difficult it is to notice the mistake, used as a proxy for detectability and perceived severity. \\
    Procedure Logic &
    Binary + Likert (1--3) &
    Whether the overall procedure becomes logically broken due to the mistake(s), together with annotator confidence. \\
    Sequence Consistency &
    Likert (1--5) &
    Whether the edited step sequence remains executable as a coherent procedure. \\
    State Change Coherence &
    Binary &
    Whether the implied world state remains coherent, without impossible preconditions or state transitions. \\
    Video Plausibility &
    Likert (1--5) &
    Whether the visual depiction of the mistake looks natural when video is available. \\
    Text-Video Grounding Consistency &
    Likert (1--5) &
    Whether the textual procedure matches what happens in the video at the episode level. \\
    \bottomrule
  \end{tabularx}
\end{table*}

To assess mistake-aware procedural traces, we separate two questions.
First, does the marked deviation constitute a consequential procedural mistake, as opposed to a permissible execution variant.
Second, if it is a mistake, is it plausible at the step level and coherent at the procedure level.
\Cref{tab:rubric_overview} summarizes the rubric dimensions and scales, following multi-criteria human evaluation practice rather than relying on a single subjective score~\cite{amidei-etal-2019-use,d7e145ce52934367931192384e305b11}.

\paragraph{Step-level criteria.} \textbf{Error Validity} is a binary gate: without recovery, the deviation would plausibly change the intended outcome, invalidate a prerequisite for later steps, or induce an incorrect intermediate state. This separates consequential mistakes from benign execution variants or incidental noise.
\textbf{Human Plausibility} measures whether a real person could naturally make this specific mistake in context.
\textbf{Confusability} measures how easy it is to miss the mistake during real-time execution; it is distinct from Human Plausibility.
\textbf{Taxonomy Fit (Error Type)} assigns each mistake to one of the five universal categories from \Cref{sec:tax} for domain-agnostic analysis across datasets and generators.
\textbf{Video Plausibility} evaluates whether the mistaken behavior looks visually natural in egocentric footage, rather than theatrical or staged.
If a \textbf{Correction} is present, it is treated as an additional step in the edited trace with an explicit semantic dependency on the triggering mistake. It is scored with the same applicable metrics, and its plausibility is interpreted relative to the mistake it addresses.

\paragraph{Procedure-level criteria.}
Procedure-level metrics assess global coherence of the full sequence of procedural steps with mistakes. \textbf{Procedure Logic} is a binary judgment of whether the entire procedure becomes logically inconsistent due to this mistake. Because this decision can be intrinsically ambiguous, it is paired with a 3-level confidence score. Let $e$ denote an annotated procedure instance and $A_e$ the set of raters who scored it.
Each rater $k\in A_e$ provides a binary decision $y_k(e)\in\{0,1\}$ indicating whether the procedure logic is broken, and a confidence weight $q_k(e)\in\{1,2,3\}$. The final per example score is computed as a confidence weighted fraction of ``logic broken'' decisions:
\begin{equation}
\label{eq:procedure-logic}
\begin{aligned}
\mathrm{PL}(e)
&=
\frac{\sum_{k\in A_e} q_k(e)\,y_k(e)}
{\sum_{k\in A_e} q_k(e)} ,
\\
&\qquad y_k(e)\in\{0,1\},\;\; q_k(e)\in\{1,2,3\}.
\end{aligned}
\end{equation}
For example, if two raters answer ``Yes'' with confidence 3 and one answers ``No'' with confidence 2, then $\mathrm{PL}(e)=6/(6+2)=0.75$. 
\textbf{Sequence Consistency Score} is a rating of whether the resulting step order remains consistent with the procedure constraints and dependencies. This captures step sequence quality beyond the binary logic decision, making a distinction between locally and completely broken reorderings.
\textbf{State Change Coherence} is a binary check that the implied world state remains consistent across the textual trace, avoiding contradictions such as objects appearing or changing identity without an action, or outcomes that become impossible under the described steps. For example, if the trace omits pouring water, but the video later shows a full coffee cup, the implied state transition is inconsistent, violating State Change Coherence.
\textbf{Text-Video Grounding Consistency (episode level)} rates alignment between the entire textual trace (including the mistake and downstream steps) and what is shown in the video. This is defined at the episode level because mismatches can accumulate after an error and can also reflect upstream step segmentation or annotation drift, not only a single step. 

\paragraph{Scalable approximations.}
Human annotations do not scale to auditing entire corpora of procedural videos, so the rubric supports learned or algorithmic proxies calibrated on a small human-labeled subset. A common approach is an LLM-based judge that predicts rubric dimensions from text for all metrics except video plausibility and video--text grounding, then calibrates its outputs against human gold labels. These proxies support large-scale screening and comparative audits; the human rubric remains the reference standard for validity and realism.

\begin{figure*}[t]
    \centering
    \includegraphics[width=0.7\linewidth]{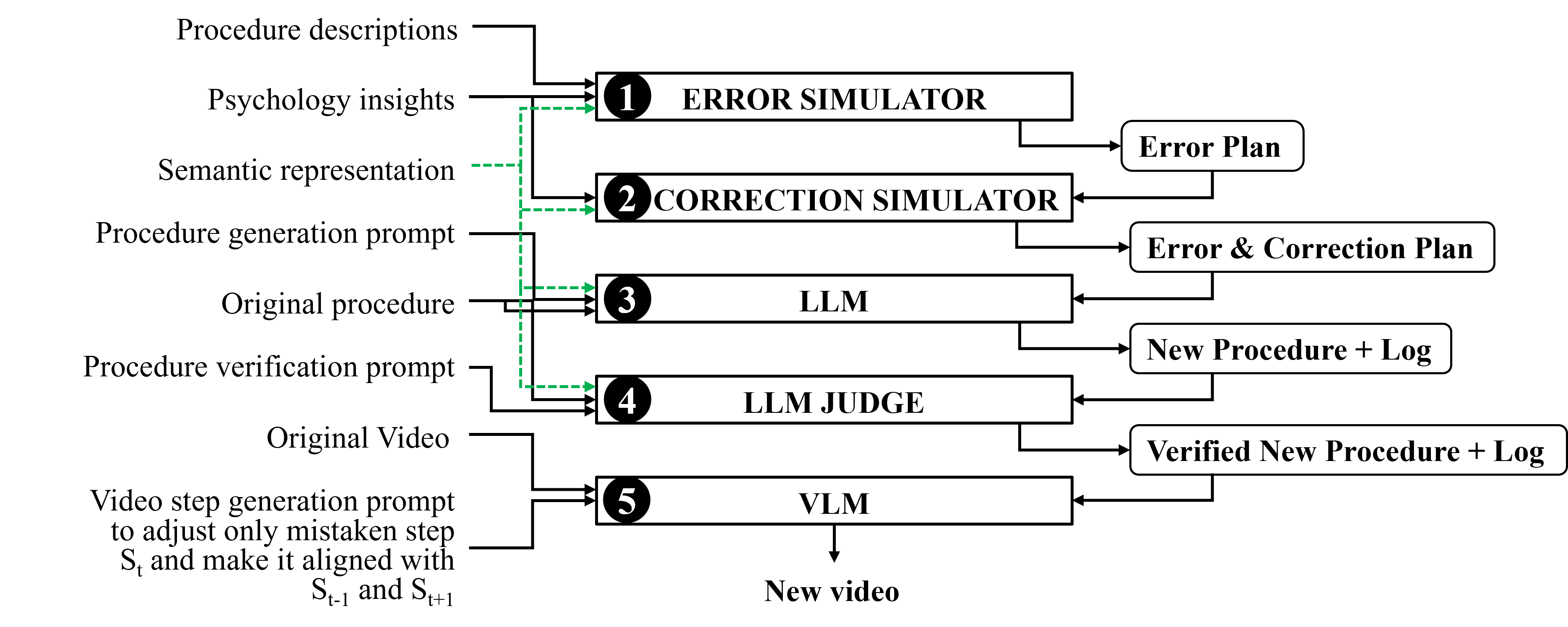}
    \caption{PIE-V pipeline overview. Clean keystep procedures are enriched by (1) an error planner (psychology-informed, constrained by step semantics and procedure phase), (2) a correction planner (recovery behavior), (3) an LLM writer (procedure rewriting with cascade consistency edits), (4) an LLM judge (coherence validation and repair; optionally multimodal), and (5) a video synthesis stage that generates new clips and smooth transitions for video plausibility. Green dashed arrows denote precomputed semantic representations for each step that condition all PIE-V modules except video generation.}
    \label{fig:piev_scheme}
\end{figure*}
\section{PIE-V algorithm}
\label{sec:method}

PIE-V augments clean keystep egocentric procedures with explicit error and recovery traces.
\Cref{fig:piev_scheme} overviews the modular pipeline: planning samples a structured error and correction program, LLM stages realize and validate a coherent textual trace, and the video stage renders edited segments so the final episode remains visually plausible.

\paragraph{Problem setup.}
We start from clean keystep procedures paired with egocentric video.
Let a reference procedure be a sequence of $T$ steps $P=(s_1,\ldots,s_T)$.
Each step $s_t$ is paired with (i) a semantic representation $a_t$ (predicate-argument structure over semantic roles) and (ii) an observed duration $d_t$ from the video segment aligned to the step.

\paragraph{Step load and phases.}
PIE-V computes phases\footnote{Phases serve as a compact control variable that separates \emph{where} errors are more likely to occur from \emph{which} error types are more likely within a step. PIE-V, in turn, encodes cognitive regularities such as peak-load sequencing failures and late post-completion omissions (details and parameter tables are provided in Sec.~\ref{sec:phases}.} using a step-load signal that combines normalized duration and a semantic-complexity proxy derived from $a_t$:
\begin{equation}
\label{eq:step-load}
\mathrm{load}(t)= w_c \cdot \widehat{\mathrm{complexity}}(t) + w_t \cdot \widehat{\mathrm{duration}}(t),
\end{equation}
where $\widehat{\mathrm{duration}}(t)$ and $\widehat{\mathrm{complexity}}(t)$ are min-max normalized within the procedure.
The complexity proxy increases with predicate count, role count, nesting depth, and the number of explicit relations in $a_t$.
We assign each step a coarse phase $\phi(t)\in\{\textsc{phase\_1},\textsc{phase\_2},\textsc{phase\_3}\}$ by splitting equally the cumulative load into thirds (early/mid/late by effort).
PIE-V outputs a mistake-aware trace consisting of a modified procedure $P'$, an error plan $E$, a correction plan $C$ (possibly empty), and an edited video episode aligned with $P'$.

\subsection{Error planner and Correction simulator}
\label{subsec:error_planner}

\subsubsection{Error plan}
PIE-V first samples an error plan $E=\{e_k\}_{k=1}^K$.
Each error event is
\[
e_k=(t_k,\tau_k,\rho_k),
\]
where $t_k\in\{1,\ldots,T\}$ is the target step index, $\tau_k$ is the error type, and $\rho_k$ stores type-specific parameters (e.g., swap partner for \textsc{T}, or mutated semantic roles for \textsc{WE}).
We use five error types $\tau\in\{\textsc{D}, \textsc{I}, \textsc{S}, \textsc{T}, \textsc{WE}\}$.

The planner is psychology-informed.
It biases error placement and type by phase $\phi(t)$ and step load $\mathrm{load}(t)$, reflecting that slips, lapses, and post-completion vulnerability vary over a procedure~\cite{Reason_1990,Norman1988ThePO,Byrne1997AWM,Byrne2006Task}.

\paragraph{Phase error-rate model (where errors occur).}
We define a phase error-rate model $r_{\phi}$ and normalize it into multipliers with mean $1$:
\begin{equation}
\label{eq:phase-mult}
m_{\phi}=\frac{r_{\phi}}{\frac{1}{3}\sum_{\phi'\in\{\textsc{phase\_1},\textsc{phase\_2},\textsc{phase\_3}\}} r_{\phi'}}.
\end{equation}
Candidate error locations are sampled with load-based weights:
\begin{equation}
\label{eq:error-step-weights}
w(t)=\bigl(\lambda+(1-\lambda)\mathrm{load}(t)\bigr)\cdot m_{\phi(t)},\qquad \lambda=0.15,
\end{equation}
under hard constraints that prevent degenerate traces (e.g., $K\le 5$ and no more than three consecutive error steps).

\paragraph{Phase-conditioned type priors (what errors occur).}
Error types are sampled from phase-conditioned priors.
Let $\pi_{\phi}$ be a phase-specific prior over the taxonomy
$\{\textsc{WE},\textsc{D},\textsc{S},\textsc{I},\textsc{T}\}$.
We sample
\begin{equation}
\label{eq:error-type}
p(\tau\mid \phi(t)) \propto \pi_{\phi(t)}(\tau)\cdot m(\tau;T,\mathrm{ess}(t)),
\end{equation}
where $m(\cdot)$ applies feasibility modifiers such as disallowing deletion if $T\le4$, limiting transposition to a local window, and biasing insertions toward non-essential steps.
We also constrain transpositions and prefer substitutions using taxonomy blocks from the underlying keystep hierarchy so that edits remain locally coherent without requiring a full world model.

\paragraph{Structural edits.}
For \textsc{D}, PIE-V removes $s_{t}$ from the trace.
For \textsc{I}, it inserts a new step near $t$ (the plan specifies insertion location and intent; the writer realizes the text).
For \textsc{T}, it swaps the order of two nearby steps within a fixed window (default window size $6$).
For \textsc{S}, it replaces the intended step with an alternative step consistent with local context and taxonomy constraints.

\paragraph{Localized edits.}
\textsc{WE} and many cases of \textsc{S} require localization inside a step. PIE-V therefore mutates role arguments inside $a_t$ rather than rewriting the entire step arbitrarily. We use a role-impact map $\omega(r)\in\{\textsc{high},\textsc{medium},\textsc{low}\}$\footnote{Role annotations are precomputed offline for the dataset step vocabulary, while the role-impact map and predicate-conditioned role priors are constructed once from the semantic representation corpus; implementation details are given in Sec.~\ref{sec:sem_rep}.}. 
For Wrong Execution, we select one (occasionally two) roles present in the step with probability proportional to an impact weight and a predicate-conditioned role prior:
\begin{equation}
\label{eq:role-sampling}
\begin{aligned}
p(r\mid a_t) \propto\;& \mathrm{ImpactWeight}(\omega(r)) \\
&\cdot \Bigl(0.2+\mathrm{Prior}\bigl(r\mid \mathrm{pred}(a_t)\bigr)\Bigr)
\end{aligned}
\end{equation}
We define error severity as the maximum impact among mutated roles.

\begin{figure}[t]
    \centering
    \includegraphics[width=0.47\textwidth]{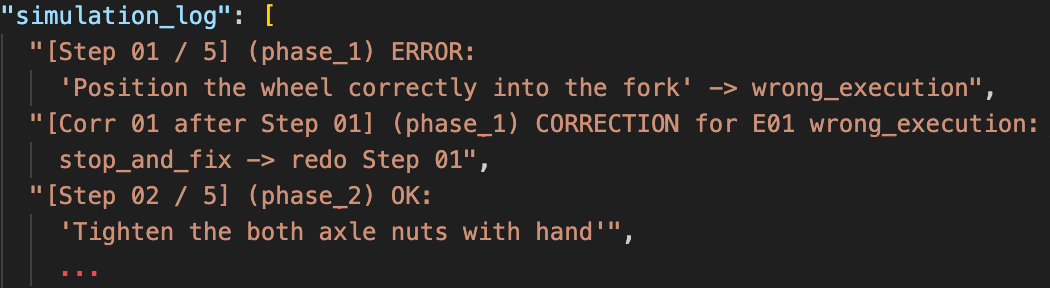}
    \caption{Example PIE-V simulation log for the Ego-Exo4D ``Install a Wheel'' task (cmu\_bike14\_4). The simulators insert a \textsc{WE} event at Step 01 (incorrect wheel positioning in the fork) and then schedule a phase-matched correction (\textsc{stop\_and\_fix} $\rightarrow$ redo Step 01) before continuing the remaining steps.}
    \label{fig:piev_sim_log}
\end{figure}

\subsubsection{Correction simulator}
\label{subsec:correction_planner}

PIE-V produces a correction plan $C=\{c_j\}_{j=1}^J$ conditioned on $E$ and procedure context (with $J$ possibly zero).
A correction event is represented as
\[
c_j=(t'_j,\kappa_j,\pi_j),
\]
where $t'_j$ is the insertion point in the edited trace, $\kappa_j$ is the correction type, and $\pi_j$ encodes the repair target (the triggering error id and the object/role to repair).

\paragraph{Detection and action priors.}
Corrections depend on whether the executor notices the error and decides to act.
We model detection with a factorized prior based on error type and phase, modulated by severity, essentiality, predicate salience, and cognitive load:
\begin{equation}
\label{eq:detect-prob}
p_{\mathrm{detect}}(e)=\mathrm{clamp}\Bigl(b(\tau,\phi)\cdot f_{\mathrm{sev}}\cdot f_{\mathrm{ess}}\cdot f_{\mathrm{pred}}\cdot f_{\mathrm{load}}\Bigr),
\end{equation}
where $b(\tau,\phi)$ is a hand-specified base detectability prior over error types and phase buckets, motivated by cognitive error recovery regularities, and
$f_{\mathrm{load}}=\max(0.70,\,1-0.25\cdot \mathrm{load}(t))$ decreases detection under high load. The full detectability tables, action priors, and latency settings are provided in Sec.~\ref{sec:corr_priors}.

Conditioned on detection, we sample a latency in steps and a correction type consistent with the triggering error, such as \textsc{stop\_and\_fix}, \textsc{redo}, \textsc{rollback\_and\_redo}, or \textsc{undo\_extra\_step}, following cognitive accounts of recovery behavior~\cite{Tamborello2013ALM}.

\Cref{fig:piev_sim_log} shows a concrete sampled trace in which \textsc{WE} at Step 01 triggers an immediate \textsc{stop\_and\_fix} correction \textsc{C} and step redo, so that the next Step\_2 remains plausible, as well as all the following steps. 

\subsection{LLM writer: coherent procedure rewriting with cascades}
\label{subsec:writer}

Given $(P,\{a_t\},E,C)$, the writer produces a rewritten procedure $P'=(s'_1,\ldots,s'_{T'})$ that instantiates planned deviations and recoveries.
A key requirement is global coherence: if an entity or attribute is edited at step $t$, later mentions must be updated consistently.
We represent this with a cascading rewrite map $M$ over entities and attributes.
For each planned local edit, we update $M$ and apply it to future steps:
\[
M \leftarrow M \cup \{v_{t,r}\mapsto \tilde{v}_{t,r}\}, \qquad
s'_{t'>t}=\mathrm{Rewrite}(s_{t'},M).
\]
The writer therefore emits both the planned error/correction steps and any necessary downstream adjustments, avoiding a common failure mode of unstructured generation where local edits silently break global procedure logic.

\subsection{LLM judge: plan compliance, coherence validation, and repair}
\label{subsec:judge}

The judge validates and repairs the writer output.
It checks three classes of constraints. \textbf{Plan compliance}: Planned error and correction events must appear at the intended locations and match the intended types, including targeted roles for localized edits. \textbf{Procedure coherence}: The rewritten trace must remain executable and logically consistent, including ordering constraints and state consistency. We treat state coherence as a predicate over implied transitions $x_{t+1}=g(x_t,s'_t)$ and reject traces that assume unavailable objects or contradict prior effects. \textbf{Recovery validity}: Corrections must address the triggering mistake and restore procedural consistency rather than introduce new contradictions.

The judge runs a bounded repair loop: it proposes minimal rewrites that preserve the plan and revalidates them.
If repeated text-only repairs fail, the judge can optionally become multimodal by attaching a small number of cached frames from the implicated steps (typically \textsc{WE} or \textsc{S}) and retrying repair with visual evidence.

\subsection{Video synthesis and stitching}
\label{subsec:video_synthesis}

PIE-V edits the egocentric episode to match $P'$ by regenerating only windows affected by planned errors or corrections and keeping other clips unchanged. For each edited window, we cache boundary anchors (end frame of the preceding step and start frame of the following step), generate a replacement clip conditioned on these anchors when supported, and splice it back, updating step timestamps. Editing is type-dependent and constrained by model duration: \textsc{WE} and \textsc{S} typically replace a full step, \textsc{I} and \textsc{C} add a short clip, \textsc{T} regenerates a local window, and \textsc{D} removes a step and inserts a brief bridge (often $<3$ s) to connect surrounding context.

\begin{table*}[t]
\centering
\caption{Aggregated rubric statistics for existing datasets and Ego-Exo4D generations. We do not report Taxonomy Fit here because it is not a scalar score/rate. Lower is better for ``Proc.Logic (Yes, \%)'' and ``State-Chg (Yes, \%)''; higher is better for the remaining reported metrics. ``--'' indicates unavailable values. The best value is highlighted only for metrics with a clear optimization direction.}
\footnotesize
\setlength{\tabcolsep}{3pt}
\resizebox{\linewidth}{!}{%
\begin{tabular}{lcccccccc}
\toprule
Dataset &
Err.Valid (Yes, \%) &
Human Pl. &
Confus. &
Proc.Logic (Yes, \%) &
Seq.Cons. &
State-Chg (Yes, \%) &
Vid.Pl. &
T--V Gr. \\
\midrule
EgoPER & 51 & 2.67 & 1.88 & 26 & 4.41 & 14 & 3.22 & 3.42 \\
EgoOops & 72 & \textbf{3.73} & 1.63 & 28 & 4.50 & 10 & \textbf{4.31} & 3.74 \\
Assembly101 & 65 & 3.69 & 2.09 & 12 & \textbf{4.82} & 4 & 4.05 & 3.22 \\
CaptainCook4D & 74 & 3.71 & 2.32 & 12 & 4.73 & \textbf{2} & 3.42 & 3.63 \\
\midrule
Ego-Exo4D-Qwen (freeform) & 57 & 3.34 & 2.04 & 36 & 4.20 & 27 & -- & -- \\
Ego-Exo4D-GPT-5.2 (freeform) & 55 & 3.08 & 1.76 & 25 & 4.18 & 7 & -- & -- \\
Ego-Exo4D-Qwen-PJ (PIE-V+Qwen2.5, Qwen3-VL-judged) & 71 & 3.09 & 1.86 & 30 & 4.39 & 10 & -- & -- \\
Ego-Exo4D-GPT-5.2-PJ (PIE-V+GPT, judged) & \textbf{89} & 3.41 & 1.76 & \textbf{6} & 4.48 & 3 & 3.54 & \textbf{3.87} \\
\bottomrule
\end{tabular}}
\label{tab:piev_stats_all}
\end{table*}
\section{Experiments}
\label{sec:exp}
\subsection{Annotations}
\label{sec:exp_annot}

\paragraph{Annotation protocol and agreement.}

We use 5 annotators (2 male, 3 female; age 20--47; mixed educational backgrounds).

Annotators evaluate samples in a paired setting: a reference execution and a mistake-aware variant with mistakes and corrections explicitly marked.
The task is to rate the quality of the indicated deviations and recoveries, not to discover them.
We monitor inter-annotator agreement using Krippendorff's $\alpha$. The details on annotators and guidelines are given in ~\ref{sec:det_annot}.

\subsection{PIE-V for Ego-Exo4D}
\label{sec:exp_pie_v_ego}

We construct a mistake-enriched benchmark from \textbf{Ego-Exo4D} by selecting 17 tasks and 50 scenarios and generating one mistake-aware variant per scenario.
The error planner injects up to five mistakes per procedure with a cap on consecutive mistakes, disallows deletions for very short procedures, and restricts transpositions to a local window.
Across the 50 scenarios, PIE-V injects 102 mistakes and 27 recovery corrections; the mistakes cover all five taxonomy types. Corrections are not generated for every mistake because recovery is sampled conditionally from detectability, action, and latency priors, as detailed in Sec.~\ref{sec:corr_priors}.

For text generation and validation in the writer/judge stages we use \textbf{GPT-5.2}~\cite{singh2025openaigpt5card}, \textbf{Qwen2.5-32B}\cite{qwen}, and the multimodal \textbf{Qwen3-VL-32B}~\cite{Qwen2VL}.
For video editing we synthesize replacement clips with \textbf{Kling-O}~\cite{klingteam2025klingomnitechnicalreport}, \textbf{Sora 2}~\cite{liu2024sorareviewbackgroundtechnology}, \textbf{Seedance 1.5 Pro}~\cite{seedance2025seedance15pronative}, \textbf{Veo 3.1}~\cite{veo31techreport2026}, and \textbf{Runway Gen-4}~\cite{runwaygen42025}.

\subsection{LLMs for Ego-Exo4D}
\label{sec:exp_llms_ego}

To assess whether unstructured generation can match structured planning, we compare PIE-V against a \textbf{freeform} baseline that rewrites a clean procedure into a mistake-aware variant directly from text instructions, without explicit phase priors or role-constrained edits.
We also evaluate a stronger baseline that adds the same validation and repair stage as PIE-V (LLM judge plus deterministic checks), isolating the effect of structured planning.
Both baselines use the same model pool as PIE-V for writer/judge: \textbf{GPT-5.2}, \textbf{Qwen2.5-32B}, and the multimodal \textbf{Qwen3-VL-32B}.
All generated traces are evaluated under the same rubric and audit protocol as the existing datasets.

\begin{table}[t]
\centering
\caption{Scale statistics for Ego-Exo4D generations under different settings.}
\footnotesize
\setlength{\tabcolsep}{3pt}
\resizebox{\linewidth}{!}{%
\begin{tabular}{lccccc}
\toprule
Setting & Total steps & Mistake steps & Mistake rate (\%) & Avg. mistakes/video \\
\midrule
Ego-Exo4D-Qwen (freeform) & 1156 & 112 & 9.69 & 2.24 \\
Ego-Exo4D-GPT-5.2 (freeform) & 1270 & 77 & 6.06 & 1.54 \\
Ego-Exo4D-Qwen-PJ & 1320 & 143 & 10.83 & 2.86 \\
Ego-Exo4D-GPT-5.2-PJ & 1323 & 141 & 10.66 & 2.82 \\
\bottomrule
\end{tabular}}
\label{tab:piev_sampling_synth}
\end{table}
\section{Results}
\label{sec:res}

\begin{table*}[t]
\centering
\caption{Krippendorff's $\alpha$ agreement summary. 
}
\footnotesize
\setlength{\tabcolsep}{4pt}
\resizebox{\textwidth}{!}{%
\begin{tabular}{lccccccccc}
\toprule
Dataset &
Err.Valid &
Human Pl. &
Confus. &
Proc.Logic &
Seq.Cons. &
State-Chg &
Taxonomy Fit &
Vid.Pl. &
T--V Gr. \\
\midrule
EgoPER &
0.912 & 0.541 & 0.368 & 0.728 & 0.628 & 0.579 & 0.759 & 0.574 & 0.662 \\
EgoOops &
0.916 & 0.592 & 0.375 & 0.836 & 0.667 & 0.600 & 0.882 & 0.579 & 0.560 \\
Assembly101 &
0.859 & 0.584 & 0.697 & 0.739 & 0.649 & 1.000 & 0.931 & 0.670 & 0.861 \\
CaptainCook4D &
0.694 & 0.758 & 0.847 & 0.621 & 0.542 & 0.584 & 0.791 & 0.550 & 0.488 \\
Ego-Exo4D-GPT-5.2-PJ (PIE-V+GPT, judged) &
0.913 & 0.489 & 0.387 & 0.672 & 0.619 & 0.696 & 0.803 & 0.630 & 0.930 \\
\bottomrule
\end{tabular}}
\label{tab:piev_alpha_full}
\end{table*}

\subsection{Audit of existing datasets}
\label{sec:audit}

We apply our rubric to four egocentric datasets with annotated mistakes:
\textbf{EgoPER}~\cite{Lee_2024_CVPR}, \textbf{EgoOops}~\cite{haneji2025egooopsdatasetmistakeaction},
\textbf{CaptainCook4D}~\cite{peddi2024captaincook4ddatasetunderstandingerrors}, and \textbf{Assembly101}~\cite{Sener2022Assembly101ALA}.
For each dataset, we randomly sample 25 videos that contain at least one annotated mistake. Table~\ref{tab:piev_sampling_existing} summarizes step and mistake density for the four audited datasets for general context.

Table~\ref{tab:piev_stats_all} summarizes rubric aggregates and reveals dataset-specific signatures relevant for procedure-level mistake reasoning: (i) how often annotated deviations are judged as consequential mistakes (Err.Valid), (ii) whether they look like errors a real person could make (Human Pl.) and whether they are easy to overlook (Confus.), and (iii) whether the resulting trace remains logically and causally coherent (Proc.Logic / Seq.Cons. / State-Chg) with aligned text and video (T--V Gr.).

Across datasets, step-level realism cues (Human Pl., Vid.Pl.) do not imply procedure-level coherence: several resources score well on plausibility while still exhibiting frequent logic or state inconsistencies under a holistic rubric.
Conversely, the higher step--mistake density summarized in Table~\ref{tab:piev_sampling_existing} often correlates with lower confusability and more staged-looking deviations, which is desirable for anomaly recognition but less representative of naturally occurring mistake-and-recovery traces.

Our audit reveals specific signatures: \textbf{EgoOops} scores strongly on Human Plausibility and Video Plausibility, suggesting mistakes tend to look behaviorally credible and visually natural. Its scenarios are specific and mistakes are mostly staged, which reduces coverage for broad everyday procedures and limits the diversity of long-horizon causal failures. \textbf{Assembly101} shows lower Text--Video Grounding Consistency. It indicates that textual step descriptions do not fully correspond to what is executed on video, which complicates episode-level tracking for multimodal models. Its completion-driven assembly protocol also reshapes the error space: genuine omissions are naturally rare, and repeated attach/detach attempts can appear insertion-like at the sequence level. \textbf{EgoPER} has comparatively low Error Validity: a substantial fraction of labeled deviations are judged as permissible variants rather than consequential mistakes. This highlights that ``mistake'' boundaries are often ambiguous in practice, and that such ambiguity can weaken supervision signals when the goal is to learn recovery-triggering errors rather than stylistic execution differences. \textbf{CaptainCook4D} combines high mistake density with lower Confusability, i.e., many deviations are easy to notice.
This profile fits segment-level anomaly recognition, but it can be less representative of naturally occurring traces where mistakes are often subtle and followed by explicit recoveries rather than frequent isolated anomalies.

Overall, these datasets were primarily designed for segment-level deviation and anomaly recognition.
Our rubric makes explicit which procedure-level properties are not directly targeted by this focus, such as coherent state transitions and episode-level text--video alignment. This reflects different design objectives rather than a flaw of the resources.

\subsection{PIE-V vs.\ LLMs}
\label{sec:piev_vs_llms}

If we compare PIE-V and freeform mistake generation, two trends stand out. First, freeform generation under-produces mistakes (\Cref{tab:piev_sampling_synth}) and produces substantially higher rates of procedure-level failures (Proc.Logic and State-Chg) despite producing locally fluent text (see \Cref{tab:piev_stats_all}).
Second, adding a judge stage without structured planning is insufficient: phase/load priors and role-constrained edits are what keep multi-error traces executable over long horizons.

A common freeform failure mode is violating implicit preconditions or dropping necessary tail steps: for instance, the model describes a deviation but leaves the step text effectively unchanged, or truncates the remaining procedure; or the resulting trace lacks a recovery step and becomes inconsistent with later state-dependent actions.

To quantify reliability of the rubric dimensions, we compute Krippendorff's $\alpha$ across annotators for each metric and dataset (\Cref{tab:piev_alpha_full}).
We expect higher agreement for crisp categorical judgments (e.g., Error Validity, Taxonomy Fit) and lower agreement for inherently subjective ratings (Human Plausibility, Confusability), where multiple interpretations of ``how a human might err'' are reasonable.

\section{Related work}
\label{sec:related}

\paragraph{Human errors, corrections, and structured procedural edits.}
PIE-V builds on cognitive accounts that distinguish \textit{slips} (execution failures) from \textit{mistakes} (planning failures)~\cite{Reason_1990, Norman1988ThePO}. We model phase-dependent vulnerability (e.g., post-completion errors)~\cite{Byrne2006Task} and elevated error rates under high cognitive load~\cite{Just1992ACT, Paas2003CognitiveLT}, and we explicitly synthesize reactive \textit{corrections} to capture human recovery behavior and memory competition effects~\cite{Tamborello2013ALM, 4e871b452b624c6c89c88c910feda1a1}. To generate realistic deviations without physically implausible corruption, we leverage semantic role labeling to localize editable arguments~\cite{palmer-etal-2005-proposition} and constrain edits by feasibility and role impact, rather than unconstrained role swaps used in misalignment generation~\cite{li2025mistakeattributionfinegrainedmistake}.

\paragraph{Procedural video benchmarks.}
Egocentric procedural datasets and mistake-focused benchmarks are rapidly expanding but remain heterogeneous in domains and taxonomies. A detailed survey and comparison are available in ~\ref{sec:datasets}. PIE-V complements these resources by providing a scalable pipeline to inject plausible, non-staged errors and recoveries across diverse scenarios, bridging domain-specific anomalies and universal procedural logic. 
\section{Conclusion}
\label{sec:concl}

Making mistakes is easy; making them \textit{correctly} is what enables reliable benchmarking. PIE-V turns mistake-aware dataset construction into a controlled, auditable pipeline for egocentric procedures. It plans phase- and load-conditioned deviations and recoveries, realizes them with constrained rewriting and validation, and synthesizes edited clips so the final episodes remain visually plausible.

PIE-V is the first method to prioritize world state when generating errors: deviations are kept only if their causal effects remain executable, consistent, and recoverable, yielding full error--correction traces rather than isolated anomalies. Human evaluation with our nine-metric rubric confirms that this structure matters: compared with freeform generation and existing resources, PIE-V more often produces consequential, procedure-coherent mistake traces with stronger plausibility cues.


\section*{Acknowledgements}
Olga Loginova thanks Amazon Alexa for supporting her research through a generous donation to Raffaella Bernardi.
{
    \small
    \bibliographystyle{ieeenat_fullname}
    \bibliography{main}
}

\setcounter{page}{1}
\maketitlesupplementary

\setcounter{section}{0}
\renewcommand\thesection{\Alph{section}}
\renewcommand\thesubsection{\thesection.\arabic{subsection}}

\setcounter{figure}{0}
\setcounter{table}{0}
\setcounter{footnote}{0}

\section{Egocentric Procedural Video Datasets (Context)}
\label{sec:datasets}

In this section, we review the main procedural video datasets, both with and without errors. Beyond a high-level comparison (Table~\ref{tab:procedural_datasets}), we describe in more detail
the datasets that are central to this work:
\textbf{EgoPER, EgoOops, Assembly101, CaptainCook4D, and Ego-Exo4D keysteps}.

Ego-Exo4D keysteps is used as source procedures for PIE-V. EgoPER, EgoOops, Assembly101, and CaptainCook4D serve as real-data references for what mistakes and corrections look like under different annotation schemes.

Most procedural datasets were created primarily for action recognition, action segmentation, key step (sequence) extraction, object interaction, or pose estimation based on visual data. Consequently, instead of a full description of the steps, the annotations may take the form of action labels.
This is typical of early datasets in the assembly domain such as MECCANO~\cite{ragusa2020meccanodatasetunderstandinghumanobject},
Assembly101~\cite{Sener2022Assembly101ALA}, ATA~\cite{Ghoddoosian2023WeaklySupervisedAS},
and IndustREAL~\cite{schoonbeek2023industrealdatasetprocedurestep}.
A shorter description of the procedure step makes it harder to recognize errors based on semantic cues.

The error annotations are heterogeneous and fragmented. The first group of general error annotations treats mistakes purely at the level of sequence validity,
i.e., whether the overall execution follows the canonical procedure, without localizing or typologizing individual erroneous steps.
In CSV~\cite{qian2022svipsequenceverificationprocedures} each video of an experiment is labeled as correct or incorrect
with respect to the entire reference protocol.
ATA~\cite{Ghoddoosian2023WeaklySupervisedAS}
focuses on detecting whether an activity sequence adheres to the expected order, emphasizing structural deviations such as deletions. The second group extends step/action annotations with per-step binary correctness labels. In HoloAssist~\cite{Wang2023HoloAssistAEA}, action segments are labeled as ``correct'' or ``mistake'', and conversational interventions are categorized (e.g., corrections, follow-ups), but the error label itself does not distinguish between procedural and executional issues. Assembly101~\cite{Sener2022Assembly101ALA}-based benchmarks used in \citet{Ding2023EveryMCA} attach a binary mistake flag to specific action segments and further distinguish structural errors such as misordering or redundant steps, along with incorrect attachment of parts. Notably, this benchmark also marks accumulating mistakes and corrective steps (detaching incorrectly attached parts) with a special label.

A smaller number of datasets introduce explicit taxonomies of both structural and execution errors, often tied to a specific domain. EgoPER~\cite{Lee_2024_CVPR} defines five error types assigned at the step level (omission, addition, modification, slip, correction). CaptainCook4D~\cite{peddi2024captaincook4ddatasetunderstandingerrors} provides a cooking-specific taxonomy
(measurement, timing, temperature, technique, missing and misordered steps).
EgoOops~\cite{haneji2025egooopsdatasetmistakeaction} adopts another multiclass execution-error taxonomy
(working with wrong objects, grasping wrong objects, correction, unintended actions, working in the wrong way, and others). CaptainCook4D and EgoOops augment each erroneous segment with a natural-language explanation of the error aligned to the procedural text (EgoOops additionally sometimes captures correction behavior in free-text explanations).

\begin{table*}[ht]
\caption{Egocentric procedural video datasets.
  \textbf{\#Steps} refers to the number of distinct action/step classes when reported; otherwise it is left as ``--''.
  For \textbf{Step annotations}, \textit{step} refers to natural step descriptions, while \textit{action label} refers to verb + object(s) phrases.
  \textit{Timestamps} mark either time or frame stamps.}
  \centering
  \scriptsize
  \setlength{\tabcolsep}{3pt}
  \begin{tabular}{p{2.2cm}llllp{2cm}p{2cm}cp{2cm}}
    \toprule
    Dataset & \#Videos & Duration [h] & \#Tasks & \#Steps & Domains & Step annotations & Mistakes & Source \\
    \midrule    MECCANO~\cite{ragusa2020meccanodatasetunderstandinghumanobject} &
      32 & -- & 1 & -- &
      toy assembly &
      action labels + timestamps &
      $\times$ & controlled lab \\
    EPIC-KITCHENS-100~\cite{damen2020epickitchensdatasetcollectionchallenges} &
      $\sim$700 & 100 & -- & $>90$k &
      cooking, kitchen activities &
      action labels + timestamps &
      $\times$ & participant recordings \\
    50 Salads~\cite{McKennaStein2012_50Salads} &
      50 & $\sim$6.4 & 1 & 17 &
      cooking &
      action labels &
      $\times$ & controlled lab \\
    EgoProceL~\cite{bansal2022viewbestviewprocedure} &
      329 & 62 & 16 & 139 &
      various, incl. cooking, assembly &
      steps + timestamps &
      $\times$ & semi-controlled participant recordings \\
    HoloAssist~\cite{Wang2023HoloAssistAEA} &
      350 & 166 & 20 & 414 &
       AR-assisted manipulations, incl. assembly &
       summary, conversations, steps + timestamps &
      \checkmark & controlled lab \\
    Assembly101~\cite{Sener2022Assembly101ALA} &
      362 & 167 & 101 & 202 &
      toy assembly &
      action labels + timestamps &
      \checkmark & controlled lab \\
      CaptainCook4D~\cite{peddi2024captaincook4ddatasetunderstandingerrors} &
      384 & 94.5 & 24 & 352 &
      cooking &
      steps + timestamps &
      \checkmark & participant recordings \\
    EgoOops~\cite{haneji2025egooopsdatasetmistakeaction} &
      50 & 6.8 & 5 & 46 &
      lab-style experiments and controlled assembly tasks &
      step + timestamp &
      \checkmark & controlled lab \\
    Ego-Exo4D (keysteps)~\cite{grauman2024egoexo4dunderstandingskilledhuman} &
  852 & 30 & 17 & 186 &
  various, incl. cooking, repair &
  step + timestamp &
  \checkmark $^\dagger$ &
  participant recordings \\
    EgoPER~\cite{Lee_2024_CVPR} &
      396 & 28 & 5 & 70 &
      cooking &
      step + timestamp &
      \checkmark & participant recordings \\
    ATA~\cite{Ghoddoosian2023WeaklySupervisedAS} &
      141 & 24.8 & 3 & 15 &
      toy assembly &
      action labels &
      \checkmark & controlled lab \\
    EPIC-Tent~\cite{Jang2019EPICTentAE} &
      24 & $>5.4$ & 1 & 38 &
      assembly &
      action labels + timestamps &
      \checkmark & participant recordings \\
    CSV~\cite{qian2022svipsequenceverificationprocedures} &
      70 & 11.1 & 14 & 106 &
      chemical experiments &
      action labels &
      \checkmark & controlled lab \\
    IndustReal~\cite{schoonbeek2023industrealdatasetprocedurestep} &
      84 & 5.8 & 2 & 75 &
      toy assembly &
      steps + timestamps &
      \checkmark & Industrial-like lab \\
    \bottomrule
  \end{tabular}
\par\vspace{2pt}
{\raggedright\footnotesize
$^\dagger$ $^\dagger$ Only 17 keysteps carry the \texttt{Mistake} label; it replaces the step description, so we use Ego-Exo4D keysteps as a clean source and inject mistakes synthetically.\par}
  \label{tab:procedural_datasets}
\end{table*}

\paragraph{Assembly101.}
In the Assembly101 annotations, each step is represented by one action class and two object classes that are being manipulated.
There are only two actions: \textit{attach} and \textit{detach}\footnote{In the original annotations there is a third rare verb class, ``position'',
which appears only together with the object ``figurine''. For the sake of simplicity it was merged into ``attach''.}.
The full object vocabulary contains 64 parts, and some of them are semantically close (for example, ``roller arm'',
``crane arm'', and ``excavator arm'' can all be seen as instances of a more general ``arm'').

The classes only record the action and the objects, so if an object is attached incorrectly (with a wrong orientation),
this can be seen only from the error-type label ``wrong orientation''.
To make the distinction clear at the text level, we converted class labels into full imperative commands using templates,
for example, ``attach the step to the chassis in the wrong way'' versus the correct ``attach the step to the chassis''.
In the original annotations, there is also no consistency between attaching part X to part Y and attaching part Y to part X.
We normalized such steps to a single canonical form. As a result, we obtain a vocabulary of 339 full step descriptions.

Some toy variants may have only a single assembly in the whole dataset (e.g., \texttt{c13c} for a single correct assembly and \texttt{b04b} for a single erroneous assembly).
Since the toy subtypes encoded by the last letter in the \texttt{toy\_id} differ only slightly, we merge them into shared type classes.

Overall, the error annotations in Assembly101 do not fully match our taxonomy, because the original annotations follow the logic of the assembly process rather than the logic of conformity to a reference procedure. For example, from the assembly point of view, detaching an incorrectly attached part can be a correct step,
but from the point of view of matching the canonical assembly, no detachment should be considered a correct step.
Similarly, re-attaching a part after an erroneous detachment may be labeled as a \textit{Correction},
but with respect to the reference procedure this step is simply correct.

From the visual point of view, actions in Assembly101 are mirror-like: a detachment is the reverse of an attachment of the same parts.
The parts in the dataset are also visually specific: they are sometimes small and visually similar to each other.
In addition, the visual referent of the same object can change as the assembly progresses.
For example, in assembly \texttt{c03f}, at the step ``attach the arm connector to the chassis'' the chassis has one appearance,
while in the next step, ``attach the body to the chassis'', the term ``chassis'' refers to two already connected parts, the arm connector and the chassis.

\paragraph{CaptainCook4D.}
A key property of the cooking domain is the need to follow precise quantities to execute a recipe successfully.
However, for visual models it is hard to see the difference between, say, ``Add 1/3 tsp salt to the pan'' (step\_id: 178)
and ``Add 1/2 tsp salt to the pan'' (step\_id: 146), and even for a human this is often unclear from a single video.
The dataset authors also note in \citet{peddi2024captaincook4ddatasetunderstandingerrors} that full recipe understanding is multimodal rather than purely visual.

Even so, some dataset steps have very similar textual descriptions but different step IDs.
For example, ``Take 1 tomato'' (step\_id: 149) versus ``Take a tomato'' (step\_id: 247),
or ``Peel 1 garlic cloves'' (step\_id: 200) versus ``Peel 1 garlic clove'' (step\_id: 14).
Such steps are visually indistinguishable and identical in their semantic representations. For each erroneous step, the modified textual description also provides corrected descriptions. However, this is not consistent.
For example, in recording \texttt{1\_33}, the first step ``Coat a 6-oz. ramekin cup with cooking spray'' is labeled with the preparation error
``Coating a large bowl instead of 6-oz ramekin cup'', yet later steps are marked as correct and described as
``Microwave the ramekin cup uncovered on high for 30 seconds'', ``Stir the ramekin cup'' and so on.
In the video the same bowl appears in all steps. Such inconsistencies in the step descriptions create discrepancies between the modalities, which are reflected in our Text--Video Grounding Consistency metric.

Some steps also share almost the same temporal segment, but receive different textual descriptions and different step IDs. For example, in \texttt{2\_28} the step ``Cut 1/8 garlic clove'' is annotated from 564.6 to 624.6 and the step ``Mince 1/8 garlic clove'' -- from 582 to 640. These steps are easy to distinguish in text but almost indistinguishable visually, adding further misalignment between the visual and textual modalities.

Given that each dataset contains less than 400 videos, these numerous inconsistencies contribute a substantial amount of noise for the models. This motivated us to include it in the list of datasets for annotators' assessment.

\paragraph{EgoPER.}
EgoPER~\cite{Lee_2024_CVPR} is an egocentric cooking dataset built around a small set of recurring recipes
(coffee, quesadilla, pinwheels, tea, oatmeal). We stratified 5 videos of each task for our annotators' assessment. 
The given dataset annotations include step-level timestamps and one of five labels of the following taxonomy:
two structural deviations (step \emph{omission} and step \emph{addition}),
two execution-level deviations (\emph{slip} and \emph{modification}), and \emph{correction}.
A distinctive aspect of EgoPER is that the annotations separate task-relevant steps from background activity:
some segments reflect incidental actions that are not part of the core procedure (e.g., reading a script on a screen),
and are marked explicitly as background rather than being forced into the step taxonomy.
This design is helpful for studying mistake detection without conflating procedural steps with incidental context.

\paragraph{EgoOops.}
EgoOops~\cite{haneji2025egooopsdatasetmistakeaction} contains 5 tasks with 10 videos per task.
While it was designed to include both mistake-free executions and scripted mistake executions,
in practice additional small deviations also appear in the ``correct'' runs (e.g., extra grasping or redundant manipulation), which makes the boundary between benign variation and mistake-like behavior particularly salient (reflected in our Error Validity metric). EgoOops gives a multi-class taxonomy of deviations (including corrections) and aligns each execution to a canonical script. However, for error segments the text often describes the \textit{deviation from the canonical step} rather than the exact action the person performed. For example, instead of restating the full step description, the annotation may specify what was wrong relative to the canonical step with a description such as ``correct errors in steps 1 and 2''. This complicates purely text-based assessment: recovering the implied correct step may require broader context and/or video grounding. Additionally, some steps are semantically dense and contain multiple predicates (e.g., ``pour ... then dip ... and squeeze ...''), which increases the structural complexity of the instruction.

\paragraph{Ego-Exo4D keysteps.}
Ego-Exo4D~\cite{grauman2024egoexo4dunderstandingskilledhuman} is a large-scale egocentric/exocentric dataset with multiple benchmarks. For PIE-V, we specifically take the split of the keystep annotations because they provide (i) step timestamps and (ii) natural-language step descriptions suitable as inputs for controlled textual rewriting. 

A practical property of the keystep annotations is that step structure can be hierarchical: a step may be a leaf or a parent over finer-grained children, and some steps are explicitly marked as non-essential (i.e., present in the video but not required for the canonical goal). In PIE-V we use all available steps when constructing clean source procedures, because this reduces the risk of deleting or transposing critical actions when injecting errors, and it preserves a faithful ``what happened'' trace. We leave for future work a more aggressive setting that injects mistakes only into essential keysteps.

\begin{table*}[t]
\centering
\caption{Audit sampling summary for the four existing mistake-aware datasets (25 videos each).}
\setlength{\tabcolsep}{4pt}
\begin{tabular}{lcccccc}
\toprule
Dataset & Videos & Total steps & Mistake steps & Mistake rate & Avg. steps/video & Avg. mistakes/video \\
\midrule
EgoPER & 25 & 358 & 134 & 37.43\% & 14.32 & 5.36 \\
EgoOops & 25 & 302 & 87 & 28.81\% & 12.08 & 3.48 \\
Assembly101 & 25 & 409 & 166 & 40.59\% & 16.36 & 6.64 \\
CaptainCook4D & 25 & 370 & 192 & 51.89\% & 14.80 & 7.68 \\
\bottomrule
\end{tabular}
\label{tab:piev_sampling_existing}
\end{table*}

\section{Details on Annotators and Annotations}
\label{sec:det_annot}

\begin{figure*}[t]
    \centering
    \includegraphics[width=\textwidth]{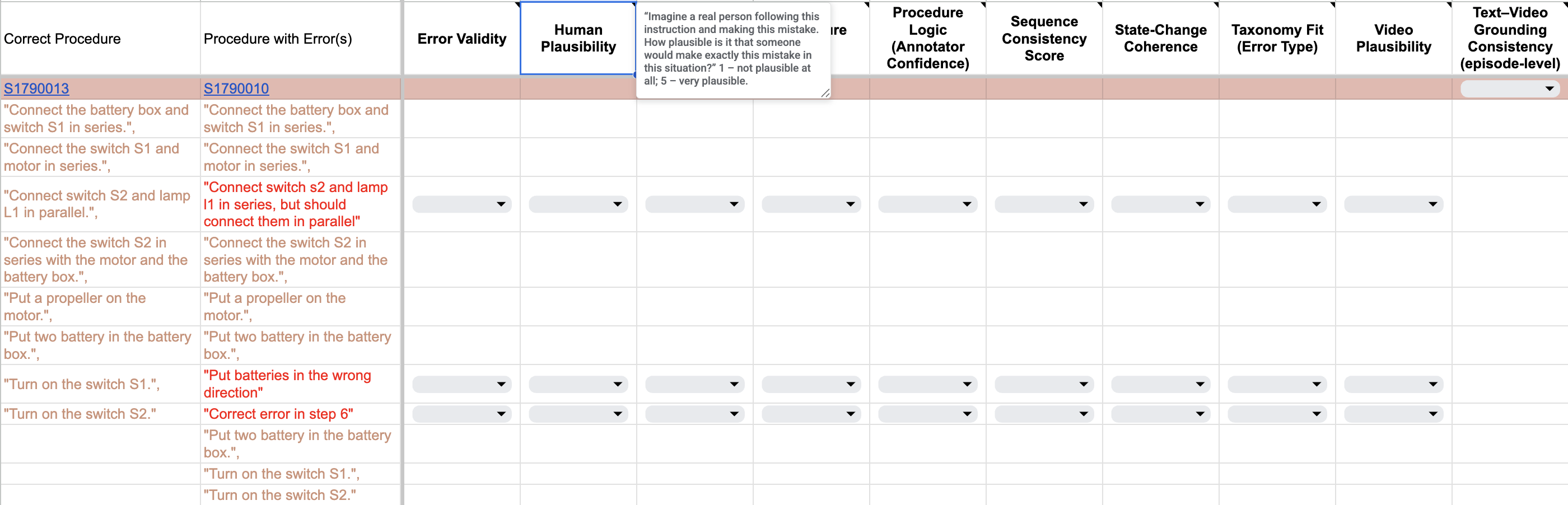}
    \caption{Annotation interface used for the rubric (example from the electronics task in EgoOops). Annotators are shown a \textbf{reference} mistake-free execution (with video) alongside a \textbf{mistake-aware} execution containing \textbf{mistakes and corrections} (highlighted in red). The goal is not to localize errors but to \textbf{rate the indicated deviations} on step-level and procedure-level criteria. Each metric includes a short in-UI explanation, and ratings are selected via drop-down menus.}
    \label{fig:annotator_interface}
\end{figure*}

\paragraph{Annotator pool and diversity of judgments.}
We use five annotators (2 male, 3 female; age 20--47) with heterogeneous backgrounds (engineering and humanities) and educational levels ranging from high school to graduate training (including two Master's degrees and one PhD).
This diversity was intentional: most rubric dimensions are designed to capture \emph{human perception} of mistake realism and coherence rather than a single objective ground truth.
In particular, Human Plausibility, Confusability, and Video Plausibility are subjective judgments by design, while Error Validity and Taxonomy Fit are expected to be more stable across annotators.

\paragraph{Annotation workflow and interface.}
Annotators evaluated samples in a paired setting with a reference (mistake-free) execution and a mistake-aware execution shown side by side.
Mistake and correction steps were explicitly highlighted in the evaluated trace; the task was to \emph{rate the indicated deviations and recoveries}, not to discover them.
The annotation interface (implemented in structured Google Sheets templates) provided metric-specific inline guidance and drop-down fields for each rating (Fig.~\ref{fig:annotator_interface}).
Reference and erroneous procedures were aligned stepwise to support direct comparison.

\paragraph{Metric design philosophy.}
With the exception of Taxonomy Fit (error category assignment), the rubric metrics were designed to capture different aspects of perceived error naturalness and procedural coherence.
Error Validity serves as a gate-like metric that distinguishes consequential procedural mistakes from non-consequential deviations; disagreement on this binary judgment captures boundary cases.
The remaining metrics decompose human judgments into plausibility (text/video), noticeability (Confusability), sequence-level coherence, and world-state consistency, rather than collapsing them into a single score.

\paragraph{Annotator instructions and scale interpretation.}
Each metric was presented with a short operational definition and anchored response options.
For example, Human Plausibility was defined as whether the described deviation looks like a mistake a real person could make in the given context, while Confusability measured how easy it would be to overlook the mistake during real-time task execution.
Procedure Logic was collected as a binary judgment (does the mistake make the textual procedure logically inconsistent as a whole) together with a 3-level confidence rating, which is later combined into a confidence-weighted aggregate score.
Sequence Consistency and Text--Video Grounding were rated on 5-point Likert scales.
State-Change Coherence was annotated as a binary judgment on whether the text implies any clear inconsistency in object identity, availability, or state transitions.

\paragraph{Pilot calibration and agreement monitoring.}
We did not expect uniformly high agreement across all metrics, because several dimensions are intentionally subjective.
Instead, we used Krippendorff's $\alpha$ as a monitoring signal and iteratively refined annotation guidelines after a pilot round, especially for metrics that produced strong outliers or systematic disagreements.
This calibration focused on clarifying metric wording and decision boundaries (e.g., procedural error vs.\ harmless variation; Substitution vs.\ Wrong Execution), not on forcing consensus.

\paragraph{Agreement summary.}
Inter-annotator agreement for all datasets and generation settings is reported in Table~\ref{tab:piev_alpha_full_1}.
Agreement is generally highest for Error Validity and Taxonomy Fit, indicating that annotators largely agree on what counts as a procedural error and how to assign taxonomy labels.
Lower agreement on Human Plausibility and Confusability reflects genuine variation in human judgments about realism and noticeability, which is an intended property of these metrics rather than a failure of the annotation protocol.

\begin{table*}[t]
\centering
\caption{Krippendorff's $\alpha$ agreement summary for all settings. ``--'' indicates values are unavailable (due to nonexistent videos).
}
\footnotesize
\setlength{\tabcolsep}{4pt}
\resizebox{\textwidth}{!}{%
\begin{tabular}{lccccccccc}
\toprule
Dataset &
Err.Valid &
Human Pl. &
Confus. &
Proc.Logic &
Seq.Cons. &
State-Chg &
Taxonomy Fit &
Vid.Pl. &
T--V Gr. \\
\midrule
EgoPER &
0.912 & 0.541 & 0.368 & 0.728 & 0.628 & 0.579 & 0.759 & 0.574 & 0.662 \\
EgoOops &
0.916 & 0.592 & 0.375 & 0.836 & 0.667 & 0.600 & 0.882 & 0.579 & 0.560 \\
Assembly101 &
0.859 & 0.584 & 0.697 & 0.739 & 0.649 & 1.000 & 0.931 & 0.670 & 0.861 \\
CaptainCook4D &
0.694 & 0.758 & 0.847 & 0.621 & 0.542 & 0.584 & 0.791 & 0.550 & 0.488 \\
Ego-Exo4D-Qwen (freeform) &
0.568 & 0.539 & 0.417 & 0.579 & 0.543 & 0.643 & 0.820 & -- & -- \\
Ego-Exo4D-GPT-5.2 (freeform) &
0.701 & 0.424 & 0.341 & 0.593 & 0.652 & 0.547 & 0.752 & -- & -- \\
Ego-Exo4D-Qwen-PJ (PIE-V+Qwen2.5, Qwen3-VL-judged) &
0.958 & 0.483 & 0.358 & 0.631 & 0.706 & 0.601 & 0.905 & -- & -- \\
Ego-Exo4D-GPT-5.2-PJ (PIE-V+GPT, judged) &
0.913 & 0.489 & 0.387 & 0.672 & 0.619 & 0.696 & 0.803 & 0.630 & 0.930 \\
\bottomrule
\end{tabular}}
\label{tab:piev_alpha_full_1}
\end{table*}

\section{Details on the Method}
\label{sec:det_method}

\subsection{Phase Modeling and Phase-Conditioned Priors}
\label{sec:phases}

PIE-V uses a three-phase abstraction to model \emph{when} errors are likely to occur and \emph{which} error types are more likely in different parts of a procedure.
The phase design follows cognitive error accounts that relate error profiles to planning demands, attentional load, routine execution, and post-completion vulnerability~\cite{Reason_1990,Norman1988ThePO,Byrne1997AWM,Byrne2006Task,Just1992ACT,Paas2003CognitiveLT}.

We model three coarse phases:
\textbf{Phase 1 (initiation)}, where plan construction and low familiarity increase attention demand;
\textbf{Phase 2 (routine execution)}, where automaticity rises but cognitive load and monotony can produce slips and sequencing failures;
and \textbf{Phase 3 (completion)}, where attention often drops and omissions become more likely due to post-completion effects.

\paragraph{Load-based phase boundaries.}
PIE-V computes step load with Eq.~\ref{eq:step-load}, forms the cumulative load over steps, and assigns \textsc{phase\_1}, \textsc{phase\_2}, and \textsc{phase\_3} by splitting the cumulative load into three equal parts.

\paragraph{Phase error-rate model (where errors occur).}
PIE-V separates phase-level risk from error type choice.
The phase error-rate model specifies relative rates
\[
r_{\textsc{phase\_1}}=0.10,\quad
r_{\textsc{phase\_2}}=0.19,\quad
r_{\textsc{phase\_3}}=0.14,
\]
which are then normalized to multipliers with mean $1$ before step-index sampling (Eq.~\ref{eq:phase-mult}).
In the current implementation, this corresponds approximately to
\[
m_{\textsc{phase\_1}}=0.698,
m_{\textsc{phase\_2}}=1.326,
m_{\textsc{phase\_3}}=0.977.
\]
Thus, the middle phase is sampled more aggressively for error placement, while total error count remains controlled separately by procedure-level risk and hard caps.

\paragraph{Phase-conditioned type priors (which errors occur).}
Conditioned on a selected step, PIE-V samples the error type from phase-specific priors over
$\{\textsc{WE},\textsc{D},\textsc{S},\textsc{I},\textsc{T}\}$.
The implementation uses unnormalized weights (in this order):
\[
\textsc{phase\_1}: [3.5,\,1.0,\,2.5,\,2.0,\,1.0],
\]
\[
\textsc{phase\_2}: [2.0,\,2.0,\,1.5,\,2.5,\,2.0],
\]
\[
\textsc{phase\_3}: [3.5,\,2.5,\,1.0,\,2.0,\,1.0].
\]
After normalization, the corresponding type probabilities are as shown in Table~\ref{tab:phase_type_priors_impl}.

\begin{table}[h]
\caption{Normalized phase-conditioned type priors used in the current PIE-V implementation (before feasibility modifiers).}
\label{tab:phase_type_priors_impl}
\centering
\small
\setlength{\tabcolsep}{4pt}
\begin{tabular}{lccccc}
\toprule
Phase & WE & D & S & I & T \\
\midrule
Phase 1 & 0.35 & 0.10 & 0.25 & 0.20 & 0.10 \\
Phase 2 & 0.20 & 0.20 & 0.15 & 0.25 & 0.20 \\
Phase 3 & 0.35 & 0.25 & 0.10 & 0.20 & 0.10 \\
\bottomrule
\end{tabular}
\end{table}

These priors are then modulated by feasibility constraints and step properties (e.g., procedure length, essentiality, transposition feasibility), as described in Sec.~\ref{subsec:error_planner}.
This factorization is intentional: phase priors encode cognitive tendencies, while feasibility modifiers prevent structurally invalid edits.

\subsection{Correction Detection and Action Priors}
\label{sec:corr_priors}

PIE-V models correction generation as a two-stage process: (i) whether an error is noticed (detection that correlates with our Confusibility metric), and (ii) whether a corrective action is taken, with what latency and repair type.
The base detectability term $b(\tau,\phi)$ in Eq.~\ref{eq:detect-prob} is a hand-specified prior over error type $\tau$ and phase bucket $\phi$, designed to reflect cognitive regularities of error noticeability and recovery behavior (e.g., salient execution failures are more detectable than subtle deviations; detection probability decreases under high cognitive load)~\cite{Reason_1990,Norman1988ThePO,Tamborello2013ALM}.

\paragraph{Base detectability table.}
We define $b(\tau,\phi)$ for each error type and phase bucket (\textsc{early}, \textsc{mid}, \textsc{late}) and then modulate it multiplicatively with severity, essentiality, predicate salience, and load factors as in Eq.~\ref{eq:detect-prob}.
These values are implementation priors (not learned parameters) and are fixed across all experiments.

\paragraph{Action, latency, and repair type.}
Conditioned on detection, PIE-V samples (i) whether the agent acts, (ii) correction latency in steps, and (iii) a correction type compatible with the triggering error (e.g., \textsc{stop\_and\_fix}, \textsc{redo}, \textsc{rollback\_and\_redo}, \textsc{undo\_extra\_step}).
This separation allows PIE-V to model both noticed-but-unfixed errors and explicit recovery traces.

\subsection{Semantic Roles}
\label{sec:sem_rep}

PIE-V uses semantic step representations as a compact structural layer between free-form step text and the error simulator.
Each step is represented as a predicate--argument expression of the form
\texttt{PREDICATE(Role: value, Role: value, ...)} with nested structures when needed (e.g., locations, purposes, temporal clauses).
These representations are used to (i) compute step complexity and phase boundaries, (ii) localize \texttt{WE} and role-level \texttt{S} edits, (iii) estimate role severity via a role-impact map, and (iv) provide predicate-conditioned role priors for selecting which arguments to mutate.

\paragraph{Source of semantic role annotations.}
For the benchmark split, semantic representations are precomputed offline for the dataset step vocabulary and cached in a JSON mapping from step id to step\_description and semantic\_representation.
We generate them with GPT-5.2 using a constrained prompt (Listing~\ref{lst:semrep_prompt}) and a strict JSON schema, in batches of step id $\rightarrow$ step description pairs, and require exact copying of input ids and step text in the output.
The generation utility also normalizes step text and builds a reverse map for robust lookup across formatting variants (e.g., punctuation differences).

\begin{lstlisting}[style=custom_style,language=,caption={Excerpt of the prompt used to generate semantic representations (SemRep) for step descriptions.},label={lst:semrep_prompt}]
You are a linguistic semantic analyzer.
For each procedural step description, generate the semantic representation in roles.

Hard constraints:
- Use compact single-line format: PREDICATE(Role: value, Role: value, ...)
- Predicates MUST be UPPERCASE.
- Prefer the role name Object (NOT Theme) for concrete manipulated entities.
- Use Agent: you unless another agent is explicitly stated.
- Prepositional phrases modifying a noun should be nested inside that noun.
- Keep entities as lowercase_with_underscores; nested structures are allowed.

Examples:
1) Insert the test swab into her nostril
INSERT(Agent: you, Object: test_swab, Destination: into(nostril(of(her))))

2) Add coffee grounds from a bowl to the filter in the French press
ADD(Agent: you, Object: coffee_grounds, Origin: bowl,
    Destination: filter(Location: in(french_press)))

3) Add cut onions to the egg in the mixing bowl
ADD(Agent: you, Object: cut(onions), Coobject: egg(Location: in(mixing_bowl)))

Return only JSON that matches the schema.
\end{lstlisting}

\paragraph{SemRep format and parsing.}
The semantic representation format is designed for controllable procedural editing rather than full semantic parsing.
Predicates are uppercase action labels, and role values are compact entity expressions with optional nesting, as shown in Listing~\ref{lst:semrep_examples} (e.g., nested \texttt{Origin}, \texttt{Destination}, \texttt{Temporal}, \texttt{Purpose}, and \texttt{Result} structures).
In the simulator, we use a shallow parser that extracts the main predicate and top-level role-value pairs from the representation string.
This is sufficient for role-targeted mutation, essentiality heuristics, and local ordering guards without introducing a full symbolic world model.

\begin{lstlisting}[style=custom_style,caption={Representative cached semantic representations (SemRep) used by PIE-V on the Ego-Exo4D split.},label={lst:semrep_examples}]
{
  "1": {
    "step_description": "Unbox covid test package",
    "semantic_representation": "UNBOX(Agent: you, Object: covid_test_package)"
  },
  "11": {
    "step_description": "Extract the test swab from her nostril",
    "semantic_representation": "EXTRACT(Agent: you, Object: test_swab, Origin: from(nostril(of(her))))"
  },
  "20": {
    "step_description": "Insert the collection swab in its pack for disposal",
    "semantic_representation": "INSERT(Agent: you, Object: collection_swab, Destination: in(pack(of(it))), Purpose: disposal)"
  },
  "21": {
    "step_description": "Cover the test vial with the lid",
    "semantic_representation": "COVER(Agent: you, Object: test_vial, Instrument: lid)"
  },
  "152": {
    "step_description": "Slowly backpedal the chain while applying the chain lube to each individual roller",
    "semantic_representation": "BACKPEDAL(Agent: you, Object: chain, Manner: slowly, Temporal: WHILE(APPLY(Agent: you, Object: chain_lube, Coobject: to(each_individual_roller(of(chain))))))"
  },
  "457": {
    "step_description": "Cut the sushi roll into smaller pieces on the cutting board",
    "semantic_representation": "CUT(Agent: you, Object: sushi_roll, Result: smaller_pieces, Location: on(cutting_board))"
  }
}
\end{lstlisting}

\paragraph{Role impact map (severity prior).}
PIE-V uses a role-impact map $\omega(r)\in\{\textsc{high},\textsc{medium},\textsc{low}\}$ to control error severity in role-level edits.
Impact labels are assigned manually based on linguistic and procedural semantics.
Roles that typically determine the manipulated entity or a critical counterpart are treated as high impact, locative and instrumental roles are typically medium impact, and manner or temporal modifiers are typically low impact.
Table~\ref{tab:sem_roles_impacts} summarizes the role inventory, counts, and impact assignments for the 50-scenario Ego-Exo4D subset used in this work.
PIE-V uses this role-impact map both for role sampling and for deriving error severity from the set of mutated roles.
Agent is excluded from mutation. 

\paragraph{Predicate-conditioned role priors.}
To avoid uniformly random role edits, PIE-V uses predicate-conditioned role priors estimated from the semantic representation corpus.
For each predicate, we aggregate the empirical frequency (or share) of roles observed with that predicate across the corpus and use the resulting distribution as $\mathrm{Prior}(r \mid \mathrm{pred})$.
At generation time, for a step with predicate $\mathrm{pred}(a_t)$ and roles present in that step, the role score is proportional to
\[
\mathrm{ImpactWeight}(\omega(r)) \cdot \bigl(0.2 + \mathrm{Prior}(r \mid \mathrm{pred}(a_t))\bigr),
\]
where the additive constant provides smoothing for rare but valid roles.
Role selection is restricted to roles present in the current step, and Agent is excluded from mutation.
PIE-V occasionally may select two roles (instead of one) for compound \textsc{WE} events.

\paragraph{Auto-extension for generated steps.}
The LLM writer can introduce new step texts that are not present in the original vocabulary.
To preserve SemRep-based validation and cascade checks, the writer pipeline supports automatic SemRep extension: unseen generated steps are resolved through a reverse text-to-id map and, if missing, are assigned new cached semantic representations via the same constrained GPT-5.2 SemRep generator. 

\begin{table}[t]
\centering
\caption{Semantic roles, counts, and impact assignments for the 50-scenario Ego-Exo4D subset used in PIE-V. Counts are given in brackets.}
\label{tab:sem_roles_impacts}
\small
\setlength{\tabcolsep}{4pt}
\renewcommand{\arraystretch}{1.08}
\begin{tabular}{@{}p{0.16\linewidth}p{0.78\linewidth}@{}}
\toprule
Impact & Roles \\
\midrule
High &
Agent (1037), Object (1027), Coobject (51) \\
Medium &
Location (288), Destination (281), Origin (259), Instrument (201), Purpose (62), Content (1) \\
Low &
Manner (28), Temporal (18), Degree (11), Path (9), Duration (5), Direction (5), Proposition (5), Result (4), Quantity (3), Theme (2), Condition (2), Criterion (1) \\
\bottomrule
\end{tabular}
\end{table}

\subsection{Cascade Edits}
\label{sec:cascade}

Cascade edits are follow-up text rewrites that preserve procedure feasibility after a planned error changes an object, tool, or local state.
They are produced in the LLM writer stage and validated in the LLM judge stage.
In the metadata, cascade edits are marked as \texttt{mod="a"} and reuse the same error id \texttt{eid} as the triggering error.
This makes the causal dependency explicit: the step is not a new mistake, but a downstream repair of textual consistency caused by an earlier mistake realization.

Cascade edits are necessary because PIE-V generates \emph{coherent traces}, not isolated error labels.
A locally valid error can make later steps impossible if references are left unchanged (e.g., a substituted object is never acquired, or a tool is removed before later use).
The writer therefore rewrites downstream steps minimally, preserving the plan while keeping the procedure executable.
The judge then checks that these adjustments are present when needed and that they remain linked to the same \texttt{eid}.

\paragraph{Example: GET-substitution propagation.}
In the Ego-Exo4D procedure \texttt{sfu\_cooking\_005\_2}, PIE-V realizes a substitution at the early GET step by replacing
\emph{``Get cucumber from the refrigerator''} with \emph{``Get bell pepper from the refrigerator''} (\texttt{eid=E01}).
This change propagates to later steps that originally depend on cucumber.
As a result, multiple downstream steps are rewritten as cascade edits (\texttt{mod="a"}, same \texttt{eid=E01}), i.e., the trace explicitly replaces cucumber-dependent steps with bell-pepper-dependent ones, including ``Wash bell pepper with water'', ``Chop bell pepper with knife on the chopping board'' (twice), and ``Add chopped bell pepper into the bowl''.
Without these cascade edits, later steps would continue referring to cucumber even though the rewritten trace acquires bell pepper instead.

\paragraph{What is and is not a cascade edit.}
A cascade edit changes a later original step only as much as needed to restore consistency with an earlier error.
It is not a primary error realization (\texttt{mod="e"}) and not a correction step (\texttt{mod="c"}).
Primary error realizations instantiate the planned mistake, while cascade edits preserve executability and semantic continuity after that mistake.

\paragraph{Writer constraints for cascades.}
The writer prompt explicitly enforces cascade behavior when an error changes inventory or object identity, including a special rule for GET-like substitutions.
A shortened excerpt is shown in Listing~\ref{lst:cascade_prompt}.

\begin{lstlisting}[style=custom_style,caption={Writer prompt excerpt enforcing cascade edits after inventory-changing mistakes.},label={lst:cascade_prompt}]
If the plan location/wording would make the sequence physically impossible, you must still realize the requested error type,
but choose the closest feasible variant near the same location (and then repair downstream references via cascade adjustments).

IMPORTANT SPECIAL CASE: GET-substitution propagation
If the planned error substitutes a GET-like step (get/take/pick/retrieve) so that you acquire Y instead of X,
then assume X is NOT available later unless it is explicitly acquired again (do NOT add new insertions unless the plan includes them).
Therefore, any later steps that refer to X should be cascade-adjusted to refer to Y:
- rewrite those downstream steps and mark them as mod="a" with the SAME eid as the GET-substitution.
This keeps the procedure executable while preserving the planned mistake.
\end{lstlisting}

\section{Details on Experiments and Results}
\label{sec:det_exp_res}

\subsection{Prompt Logic and Structured Output Contracts}
\label{sec:prompts}

We do not reproduce the full LLM prompts here because they are long and implementation-specific. The full writer and judge prompts are available in the released codebase.
Here, we summarize the core constraints they enforce and the structured output contracts that are required by the pipeline.

The writer prompt receives the original procedure, an error and correction plan, semantic representations of original steps, and ordering constraints.
Its main objective is to realize the planned mistakes while keeping the rewritten procedure physically feasible and executable.
The prompt explicitly requires structured JSON output with a rewritten step list (\texttt{final\_steps}) and a timeline mapping (\texttt{meta}) that marks unchanged steps, primary error realizations, cascade edits, insertions, deletions, corrections, and transposition pairs.

The judge prompt validates plan compliance and procedural coherence and proposes minimal repairs when needed.
In particular, it checks that each planned error and correction id is realized, that transpositions are encoded via the required \texttt{ms}/\texttt{mt} pair, and that downstream references are repaired through cascade edits when an earlier error changes object or tool availability.
These prompt-level constraints are combined with deterministic checks in the pipeline, so acceptance depends on both LLM reasoning and rule-based validation.
Representative prompt fragments for cascade constraints and video style locking are shown in Listings~\ref{lst:cascade_prompt} and~\ref{lst:style_lock_prompt}.

For comparison, Listing~\ref{lst:freeform_prompt} shows the freeform baseline prompt used to generate mistake-aware procedures without structured planning.

\begin{lstlisting}[style=custom_style,caption={Freeform LLM baseline prompt for direct mistake-aware procedure rewriting (without PIE-V planning priors).},label={lst:freeform_prompt}]
instructions = (
    f"### ROLE: Procedure Editor for '{scenario}'\n"
    f"You will receive a step-by-step procedure.\n"
    f"Your task is to produce an edited final procedure that contains 1 to 3 plausible human mistakes.\n"
    f"You may also add 0 to 2 explicit correction steps.\n\n"
    f"### ERROR TYPES (HIGH-LEVEL)\n"
    f"- wrong_execution (mod='we'): Keep the same general goal, but execute it slightly wrong (wrong amount, messy action, wrong orientation).\n"
    f"- substitution (mod='s'): Replace the step with a different plausible action caused by confusion.\n"
    f"  Do not copy a later step verbatim.\n"
    f"- insertion (mod='i'): Insert one extra plausible step (unnecessary repetition, extra cleaning, checking, etc.).\n"
    f"- deletion: Remove the step completely (do not mention it was skipped). Add it to 'del'.\n"
    f"- transposition: Swap two steps.\n"
    f"  Use mod='ms' for the moved SOURCE step and mod='mt' for the moved TARGET step (both verbatim text, only order changes).\n\n"
    f"### HARD CONSTRAINTS\n"
    f"1) VERBATIM PRESERVATION: Keep most steps unchanged unless directly edited.\n"
    f"2) IMPERATIVE STYLE: Use direct imperative commands. No story.\n"
    f"3) SOURCE INDEX RANGE: Every meta source_idx MUST be a valid original index in [0, {len(steps)-1}]. Never use -1.\n"
    f"4) PHYSICAL PLAUSIBILITY: Do not use tools/ingredients/objects before they appear earlier in the procedure.\n"
    f"5) METADATA ALIGNMENT: 'final_steps' and 'meta' must have the exact same length.\n\n"
    f"6) UNCHANGED MEANS VERBATIM: If mod='u', the final step text MUST match the original step text exactly.\n"
    f"7) TRANSPOSITION RULE: If using transposition, error_id must be non-null and exactly TWO steps must share that error_id:\n"
    f"   - one with mod='ms' and one with mod='mt'.\n\n"
    f"8) NO FAKE ERRORS: If mod in ['we','s'], the text MUST be meaningfully different from the original step at source_idx.\n"
    f"9) MOVE IS VERBATIM: If mod in ['ms','mt'], the text MUST be identical to the original step at source_idx (only moved).\n"
    f"10) INSERTION IS NEW: If mod='i', the inserted text MUST NOT be identical to the anchor step at source_idx.\n"
    f"11) CORRECTION NEEDS ID: If mod='c', correction_id must be non-null like 'C01' and the text must be new.\n"
    f"12) ERROR IDS: For mod in ['we','s','i','ms','mt'] use error_id like 'E01'. For deletions also.\n"
    f"### OUTPUT FORMAT (STRICT JSON ONLY)\n"
    f"- final_steps: list[str]\n"
    f"- meta: list[list], one per final step: [source_idx, mod, error_id, correction_id]\n"
    f"  mod_type: 'u'(unchanged), 'we'(wrong_execution), 's'(substitution), 'ms'(moved_source), 'mt'(moved_target), 'c'(correction), 'i'(insertion)\n"
    f"- del: list[list] for deleted steps: [source_idx, error_id]\n\n"
    f"For every deletion, error_id must be a string like 'E01' (never null).\n"
    f"Return ONLY a single JSON object. No markdown. No extra text.\n"
    f"Do NOT repeat the prompt. Output ONLY JSON.\n"
)
\end{lstlisting}

\subsection{Annotation Data Processing and Aggregation}
\label{sec:ann_agg}

This subsection describes how raw annotator responses are converted into the aggregate statistics reported in Tables~\ref{tab:piev_stats_all} and~\ref{tab:piev_alpha_full}.

\paragraph{Metric types.}
Our rubric mixes categorical judgments (e.g., Error Validity, Procedure Logic, State-Change Coherence, Taxonomy Fit) and Likert-scale ratings (e.g., Human Plausibility, Confusability, Sequence Consistency, Video Plausibility, Text--Video Grounding).
This is intentional, because the rubric is designed to capture both relatively stable categorizations and subjective human judgments about realism and noticeability.

\paragraph{Error Validity.}
Error Validity is annotated as a binary judgment (Yes/No), namely whether the highlighted deviation is a consequential procedural error under our definition.
In the main-paper tables, we report the percentage of ``Yes'' judgments aggregated at the sample level.
Disagreements on this metric therefore reflect ambiguity in the generated behavior (or dataset annotation), not a multi-level scoring scheme.
In practice, disagreement on this binary metric often comes from freeform outputs that are lexically marked as ``accidental'' but have weak procedural consequences.
We provide representative examples in Sec.~\ref{sec:qual_failures}.

\paragraph{Procedure Logic with confidence weighting.}
Procedure Logic is annotated as a binary judgment together with confidence; aggregation follows the confidence-weighted formulation defined in Eq.~\ref{eq:procedure-logic}.

\paragraph{Likert-scale aggregation.}
For Human Plausibility, Confusability, Sequence Consistency, Video Plausibility, and Text--Video Grounding, we report the arithmetic mean over annotator ratings.

\paragraph{Agreement.}
We compute Krippendorff's $\alpha$ separately for each metric and setting.
Higher agreement is expected for Error Validity and Taxonomy Fit, while lower agreement on Human Plausibility and Confusability reflects genuine variation in human judgments about naturalness and noticeability.

\subsection{Additional Breakdowns for Ego-Exo4D Generations}
\label{sec:egoexo_breakdowns}

Table~\ref{tab:piev_sampling_synth} reports scale statistics for the four Ego-Exo4D generation settings.
Two patterns are consistent across the audited outputs.
First, freeform generation under-produces mistakes relative to PIE-V+Judge settings, with fewer mistake steps and a lower average number of mistakes per video.
Second, even when freeform outputs are fluent at the sentence level, they more often fail to realize clearly consequential deviations or to preserve long-horizon procedural consistency, which is reflected in the rubric aggregates in Table~\ref{tab:piev_stats_all}.

The freeform models also show different error-type biases.
For Qwen freeform, the generated distribution is strongly skewed toward deletions (Deletion 31, Substitution 12, Transposition 5, Wrong Execution 5, Insertion 2, Correction 1), and qualitative inspection shows many weak or under-realized edits.
For GPT freeform, the distribution is more concentrated on Wrong Execution and Substitution (Wrong Execution 27, Substitution 13, Transposition 8, Insertion 2, Correction 4, Deletion 0), which often produces locally plausible deviations but still under-produces explicit recovery behavior relative to PIE-V.
These tendencies motivate the use of explicit planning, cascade edits, and judge-side validation in PIE-V.

\subsection{Qualitative Failure Cases of Freeform Baselines}
\label{sec:qual_failures}

\paragraph{Binary Error Validity disagreement from weakly consequential freeform insertion (\texttt{indiana\_bike\_03\_1}).}
A recurring source of disagreement on Error Validity in freeform generations is that the LLM marks an event as accidental without introducing a clearly consequential procedural failure.
For example, in a bike-chain cleaning procedure, GPT inserts
\emph{``Accidentally knock the chain lube bottle over on the floor while reaching for it''}
between
\emph{``Hold the toothbrushes to the chain as you backpedal with your other hand''}
and
\emph{``Get the chain lube from the floor''}.
Some annotators judge this as a procedural error, while others judge it as a harmless deviation because the procedure remains executable and the intended outcome is not meaningfully affected.
This is precisely the kind of boundary case that Error Validity is designed to expose.

\paragraph{Qwen repetition instead of a meaningful substitution (\texttt{nus\_covidtest\_15\_2}).}
In a COVID-test procedure, a planned substitution near the disposal and waiting stage should replace the step
\emph{``Arrange test materials in the plastic bag for disposal''}.
Qwen, even with judge, realizes the deviation by duplicating the previous waiting step, producing two consecutive occurrences of \emph{``Wait for a few minutes''}.
This keeps the text locally fluent but weakens procedural specificity and does not create a clear, interpretable mistake mechanism.

By contrast, GPT produces a more consequential and traceable deviation in the same region, for example \emph{``Dispose the test plate into the plastic bag''}, followed by a compensating step \emph{``Takes the test plate from the plastic bag to check the test plate for the results''}.
Although still imperfect, this sequence preserves a clearer error and recovery interpretation for benchmarking.

More broadly, freeform outputs often use lexical markers such as \emph{``Accidentally''} or \emph{``Stop and ...''} while describing behavior that is only weakly harmful, visually subtle, or procedurally neutral.
This is one reason Error Validity remains an informative metric for freeform baselines: it measures whether the generated deviation is perceived as a consequential procedural error, not whether the text merely sounds like one.

\subsection{From LLM Step Text to Video Generation Prompts}
\label{sec:video_prompt_compilation}

A practical scalability bottleneck of PIE-V is the conversion of LLM-generated step text into video-generation prompts.
Writer and judge outputs are optimized for procedural coherence and annotation traceability, not for direct video synthesis.
As a result, many generated step descriptions require manual prompt compilation to make the intended visual event physically explicit and compatible with a specific video model.

This issue is especially visible in freeform generations.
LLMs often produce text that is linguistically marked as an error, for example with words such as \emph{``accidentally''}, while the described behavior is visually subtle, weakly consequential, or underspecified for video generation.
Similarly, correction steps may be described as meta-actions, for example \emph{``Stop and fix...''}, that require rewriting into concrete visible behavior before synthesis.

In our workflow, each edited segment is therefore paired with a model-specific prompt that specifies observable actions, object identity, camera constraints, and scene continuity.
This prompt compilation step is currently the main scalability bottleneck of the video stage in PIE-V.

\begin{lstlisting}[style=custom_style,caption={Example style-lock prompt used for egocentric video generation (bike-repair scenario).},label={lst:style_lock_prompt}]
PROMPT_STYLE_LOCK = (
    "Egocentric head-mounted camera, fixed POV, same fisheye lens and same dark circular vignette. "
    "Same bike repair workshop, same bike wheel and tools, same hands and body, same lighting. "
    "No new objects, no swaps, no text overlays, no reframing, no cut."
)
\end{lstlisting}

For Ego-Exo4D edits, a short style-lock prefix is often sufficient to preserve egocentric viewpoint and scene identity across regenerated clips (Listing~\ref{lst:style_lock_prompt}).

\subsection{Video Model Constraints and Editing Windows}
\label{sec:video_model_constraints}

Video generation models differ in conditioning interface and clip-length limits.
Some support text-only generation, while others require boundary anchors, for example start and end frames, or support only specific durations.
Because of these constraints, long procedural steps cannot always be replaced in a single pass and often require windowing, bridge clips, and stitching.

In PIE-V, edited windows are selected around the targeted mistake or correction, while unchanged parts of the episode are preserved from the original video.
When a full-step replacement is not feasible, we generate shorter clips and reconnect them with boundary-aware splicing and transition smoothing.
This is particularly important for egocentric video, where continuity of hands, tools, and camera viewpoint strongly affects plausibility.

Different video models also require different conditioning inputs, including text-only mode, boundary-frame mode, and model-specific editing interfaces, which prevents a single universal editing script.
We also tested prompt-based editing on non-egocentric videos, but support is uneven across models, especially for clips with clearly visible people, which makes the current workflow more reliable on egocentric footage.

\subsection{Limitations and Future Directions}
\label{sec:limitations}

PIE-V is intended as a controlled framework for constructing and auditing mistake-aware procedural traces, and the current version should be understood as a first step rather than a final fully scaled benchmark.

\paragraph{Benchmark scale and coverage.}
The current benchmark remains modest in scale: it is built from 50 Ego-Exo4D scenarios and contains 102 injected mistakes and 27 recovery corrections. This is sufficient for an initial controlled study, but it does not yet support strong claims of exhaustive coverage over mistake diversity, recovery patterns, or domain variation. In particular, some error types, correction strategies, and long-horizon dependencies are still underrepresented. Expanding the benchmark with more tasks, multiple variants per scenario, and broader procedural domains is a natural next step.

\paragraph{Downstream utility is not yet established.}
The present paper evaluates PIE-V primarily through human judgment and comparative auditing. While this is appropriate for validating plausibility and coherence, we do not yet show that PIE-V data improves downstream models for mistake detection, correction prediction, or post-completion verification. Demonstrating such gains through controlled training and transfer experiments is an important direction for future work.

\paragraph{The video stage is not yet fully automated, but it already changes the cost profile of mistake-aware data construction.}
The textual planning and rewriting components are substantially more scalable than the current video-editing stage. In practice, converting rewritten procedural steps into high-quality video-generation prompts still requires model-specific prompt compilation and manual iteration. This bottleneck is compounded by heterogeneous video-model interfaces, conditioning requirements, and clip-length limits, which prevent a single universal editing pipeline.

At the same time, PIE-V operates under a fundamentally different data-construction paradigm from existing mistake-aware procedural video datasets, including the ones we audit, which rely on newly recorded human executions of erroneous procedures. Such collection typically requires participant time, repeated performances, annotation effort, and often specialized capture setups. By contrast, PIE-V starts from existing clean procedural videos and edits only targeted segments. In this sense, even in its current partially manual form, PIE-V already offers a practical and resource-efficient alternative to full re-recording, reducing both collection cost and human effort while preserving control and auditability.

To our knowledge, PIE-V is also among the first mistake-aware procedural benchmark constructions to rely on prompt-based editing of existing video segments rather than new recordings of erroneous executions. We therefore view the current pipeline not only as a benchmark-generation method, but also as an early demonstration of a different and potentially much cheaper way to build mistake-aware procedural video resources.

We also expect the scalability of this stage to improve as prompt-based and instruction-guided video editing models continue to advance~\cite{ku2024anyv2v,li2025egoedit,mai2025easyv2v}. Recent progress in general video-to-video editing, instruction-based video editing, and emerging egocentric video editing suggests that more automated and temporally consistent editing pipelines may substantially reduce the need for manual prompt engineering. However, adapting such models to controlled procedural mistake construction remains a separate research problem.

\paragraph{Video-side evaluation is necessarily selective.}
A limitation of the current study is that Video Plausibility and Text--Video Grounding are not reported for every generation setting. Producing and auditing fully edited videos is substantially more resource-intensive than text-only evaluation, since each setting requires video generation, clip selection, temporal stitching, and additional human assessment of the resulting outputs. For this reason, we adopted a staged evaluation design: generation settings were first compared at the textual/procedural level, and only the strongest configuration was carried forward to the full video stage. We view this as an intentional and pragmatic use of limited computational and annotation resources rather than as an arbitrary omission, because scaling clearly weaker text-generation settings to costly video realization would add substantial expense with limited scientific value. A broader cross-setting video evaluation remains an important direction for future work, but it would require substantially greater compute, annotation time, and human effort.

\paragraph{Dependence on source data and external models.}
The current benchmark is derived from Ego-Exo4D keysteps and therefore inherits both the strengths and the limitations of that source representation. More broadly, PIE-V also depends on rapidly evolving LLM and video-generation models whose behavior, interfaces, and output quality may change over time. For this reason, the current results should be interpreted as evidence for the usefulness of the PIE-V design principles and evaluation protocol, rather than as a claim that one fixed generated benchmark is final or universal.

\paragraph{Generated mistakes remain approximations of human behavior.}
Even when errors are psychology-informed, role-constrained, and judged coherent by annotators, generated traces remain approximations rather than direct observations of naturally occurring human mistakes. They may miss social context, tacit goals, embodied variability, and opportunistic recovery strategies that arise in real-world execution. We therefore view PIE-V as complementary to naturally observed mistake datasets rather than a replacement for them.

\end{document}